\documentclass{article}

\usepackage{bnaic}
\usepackage{amssymb}
\usepackage{oz}

\newtheorem{Definition}{Definition}[section]
\newtheorem{Assumption}[Definition]{Assumption}
\newtheorem{Conjecture}[Definition]{Conjecture}
\newtheorem{Constraint}[Definition]{Constraint}
\newtheorem{Convention}[Definition]{Convention}
\newtheorem{Corollary}[Definition]{Corollary}
\newtheorem{Example}[Definition]{Example}
\newtheorem{Lemma}[Definition]{Lemma}
\newtheorem{Notation}[Definition]{Notation}
\newtheorem{Remark}[Definition]{Remark}
\newtheorem{Theorem}[Definition]{Theorem}

\newenvironment{definition}{\begin{Definition}\rm}{\end{Definition}}

\newenvironment{lemma}{\begin{Lemma}\rm}{\end{Lemma}}

\newenvironment{theorem}{\begin{Theorem}\rm}{\end{Theorem}}

\newcommand{\transru}[2]
{\shortstack{$\underline{\;\;{#1}\;\;}$ \\ $\;\;{#2}\;\;$}}
\newcommand{\transro}[2]
{\shortstack{$\;\;{#1}\;\;$ \\ $\overline{\;\;{#2}\;\;}$}}
\newcommand{\transroax}[2]
{\shortstack{$\;\;{#1}\;\;$ \\ $\;\;{#2}\;\;$}}

\newcommand{\Bact}{{\sf Bact}}		

\newcommand{\La}{{\cal{L}}}		
\newcommand{\LG}{{\sf Goal}}		
\newcommand{\LR}{{\sf Rule}}		

\newcommand{\bbase}{\Sigma}		
\newcommand{\gbase}{\Gamma}		
\newcommand{\mstate}{\langle\bbase,\gbase\rangle}
\newcommand{\mstateprime}{\langle\bbase',\gbase'\rangle}

\newcommand{\update}{{\cal M}}

\newcommand{\ac}{{\sf a}}		

\newcommand{\myins}{{\sf ins}}
\newcommand{\mydel}{{\sf del}}

\newcommand{\adopt}{{\sf adopt}}
\newcommand{\drop}{{\sf drop}}

\def\true{{\sf true}}
\def\false{{\sf false}}
\newcommand{\impl}{\rightarrow}		
\newcommand{\limpl}{\leftarrow}		
\newcommand{\bimpl}{\leftrightarrow}	

\newcommand{\bel}{{\sf B}}		
\newcommand{\goal}{{\sf G}}

\newcommand{\initc}{{\bf init}}

\newcommand{\eventual}{{\diamond}}
\newcommand{\until}{\;{\bf until}\;}
\newcommand{\unless}{\;{\bf unless}\;}
\newcommand{\ensures}{\;{\bf ensures}\;}

\newcommand{\Hoare}[3]{{\{#1\}\;#2\;\{#3\}}}
\newcommand{\HoareV}[3]
{\shortstack[l]{$\;\{#1\}\;$ \\ $\;\;\;\;{#2}\;$ \\ $\;\{#3\}\;$}}
\newcommand{\precond}{\rho}
\newcommand{\postcond}{\sigma}

\newcommand{\Impl}{\Rightarrow}		

\newcommand{\lra}{\longrightarrow}

\newcommand{\wlp}{\mbox{\em wlp}}
\newcommand{\enabled}{\mbox{\em enabled}}

\newcommand{\Trans}{{\cal T}}

\newcommand{\arr}{\rightarrow}		
\newcommand{\cmodels}{\models_C}
\newcommand{\mmodels}{\models_M}
\newcommand{\cvdash}{\vdash_C}
\newcommand{\mvdash}{\vdash_M}
\newcommand{\hmodels}{\models_H}
\newcommand{\hvdash}{\vdash_H}

\newcommand{\weg}[1]{}

\title{Agent Programming with Declarative Goals}
\author{F.S. de Boer \affila\affilb \ \ K.V. Hindriks \affila\ \   
 W. van der Hoek \affilc\ \ 
J.-J.Ch. Meyer \affila}
\date{\affila\ Institute of Information \& Computing Sciences, Utrecht University, The Netherlands \\
P.O. Box 80 089, 3508 TB Utrecht $\{${\tt koenh,frankb,jj}$\}$@{\tt cs.uu.nl}\\
\affilb\ National Research Institute for Mathematics and Computer Science (CWI),
Amsterdam, The Netherlands \\
\affilc\ Department of Computer Science, University of Liverpool, United
Kingdom $\{${\tt wiebe}$\}$@{\tt csc.liv.ac.uk}
}

\def\sc{\em}
\begin{document}
\newcounter{mpar}
\ttl
\thispagestyle{empty}

\begin{abstract}
A long and lasting problem in agent research has been to close the gap
between agent logics and agent programming frameworks. The main reason
for this problem of establishing a link between agent logics and agent
programming frameworks is identified and explained by the fact that
agent programming frameworks have not incorporated the concept of a
{\em declarative goal}. Instead, such frameworks have focused mainly
on plans or {\em goals-to-do} instead of the end goals to be realised
which are also called {\em goals-to-be}. In this paper, a new programming
language called GOAL is introduced which incorporates such declarative
goals. The notion of a {\em commitment strategy} - one of the main
theoretical insights due to agent logics, which explains the relation
between beliefs and goals - is used to construct a computational semantics
for GOAL. Finally, a proof theory for proving properties of GOAL agents is
introduced. Thus, we offer a complete theory of agent
programming in the sense that our theory provides both for a programming
framework and a programming logic for such agents.
An example program is proven correct by using this
programming logic.
\end{abstract}

\section{Goal-Oriented Agent Programming}
Agent technology has come more and more into the limelight of computer
science. Intelligent agents have not only become one of the central topics
of artificial intelligence (nowadays sometimes even defined as ``the study
of agents",~\cite{RN95}), but also mainstream computer science, especially
software engineering, has taken up agent-oriented programming as a new
and exciting paradigm to investigate, while industries experiment with the
use of it on a large scale, witness the results reported in conferences like
Autonomous Agents (e.g.~\cite{AA97}) and books like e.g.~\cite{JW97}.

Although the definition of an agent is subject to controversy, many researchers
view it as a software (or hardware) entity that displays some form of
{\em autonomy}, in the sense that an agent is both {\em reactive} (responding
to its environment) and {\em pro-active} (taking initiative, independent of
a user). Often this aspect of autonomy is translated to agents having a {\em
mental state} comprising (at least) {\em beliefs} on the environment and
{\em goals} that are to be achieved (\cite{Wool95}).

In the early days of agent research, an attempt was made to make the concept
of agents more precise by means of {\em logical systems}. This effort
resulted in a number of - mainly - modal logics for the specification of
agents which formally defined notions like {\em belief}, {\em goal},
{\em intention}, etc. associated with agents \cite{Rao96,Lind96,Coh90,Coh95}.
The relation of these logics with more practical approaches remains unclear,
however, to this day. Several efforts to bridge this gap have been attempted.
In particular, a number of {\em agent programming languages} have been
developed to bridge the gap between theory and practice \cite{Rao96a,Hin97}.
These languages show a clear family resemblance with one of the first agent
programming languages Agent-0 \cite{Sho93,Hin99a}, and also with the language
ConGolog \cite{Gia99,Hin9807,Hin00}.

These programming languages define agents in terms of their corresponding
beliefs, goals, plans and capabilities. Although they define similar
notions as in the logical approaches, there is one notable difference. In
logical approaches, a goal is a {\em declarative} concept, 
(also called a {\em goal-to-be}), whereas in
the cited programming languages goals are defined as sequences of
actions or {\em plans} (or {\em goals-to-do}). 
The terminology used differs from case to case.
However, whether they are called commitments (Agent-0), intentions
(AgentSpeak \cite{Rao96a}), or goals (3APL \cite{Hin99b}) makes little
difference: all these notions
are structures built from {\em actions} and therefore similar in nature
to {\em plans}. With respect to ConGolog, a more traditional computer
science perspective is adopted, and the corresponding structures are simply
called programs. The PLACA language \cite{Tho93}, a successor of AGENT0,
also focuses more on extending AGENT0 to a language with complex planning
structures (which are not part of the programming language itself!) than
on providing a clear theory of declarative goals of agents as part of a
programming language and in this respect is similar to AgentSpeak and 3APL.
The type of goal included in these languages may also be called a
{\em goal-to-do} and provides for a kind of {\em procedural} perspective
on goals.

In contrast, a {\em declarative} perspective on goals in
agent languages is still missing. Because of this mismatch it has not
been possible so far to use modal logics which include both belief and goal
modalities for the specification and verification of programs written in
such agent languages and it has been impossible to close the gap
between agent logics and programming frameworks so far. The value of
adding declarative goals to agent programming lies both in the fact that
it offers a new abstraction mechanism as well as that agent programs with
declarative goals more closely approximate the intuitive concept of an
intelligent agent. To fully realise the potential of the notion of an
intelligent agent, a declarative notion of a goal, therefore, should also be
incorporated into agent programming languages. 

In this paper, we
introduce the agent programming language GOAL (for
Goal-Oriented Agent Language), which takes the declarative
concept of a goal seriously and which provides a concrete proposal to bridge
the gap between theory and practice. GOAL is inspired
in particular by the language UNITY
designed by Chandy and Misra \cite{Cha88}, be it that GOAL incorporates
complex agent notions.
We offer a complete theory of agent
programming in the sense that our theory provides both for a programming
framework and a programming logic for such agents. In contrast with other
attempts \cite{Sho93,Wob99} to bridge the gap, our programming language and
programming logic are related by means of a  formal semantics. Only by
providing such a formal relation it is possible to make sure that statements
proven in the logic concern properties of the agent.

\section{The Programming Language GOAL}
In this section, we introduce the programming language GOAL. As 
mentioned in the previous section, GOAL is influenced by
by work in concurrent programming, in particular by the language UNITY
(\cite{Cha88}). The basic idea is that a
set of actions which execute in parallel constitutes a program. However,
whereas UNITY is a language based on assignment to variables, the language
GOAL is an agent-oriented programming language that incorporates more
complex notions such as belief, goal, and agent capabilities which operate
on high-level information instead of simple values.

\subsection{Mental States}
As in most agent programming languages, GOAL agents select actions on the
basis of their current mental state. A mental state consists of the beliefs
and goals of the agent. However, in contrast to most agent languages, GOAL
incorporates a {\em declarative} notion of a goal that is used by the agent
to decide what to do. Both the beliefs and the goals are drawn from one and
the same logical language, $\La$, with associated consequence relation
$\cmodels$. In this paper, $\La$ is a propositional language, and one
may think about $\cmodels$ as `classical consequence'. 
In general however, the language $\La$ may also 
be conceived as an arbitrary
{\em constraint system}, allowing one to combine tokens
(predicates over a given universe)
using the operator $\land$ (to accumulate pieces of information) and
$\exists_x$ (to hide information) to represent constraints
over the universe of discourse (Cf.~\cite{saraswat91}).
In such a setting, one often assumes the presence of a
constraint solver that tests $\Gamma \cmodels \varphi$,
i.e., whether information $\Gamma$ entails $\varphi$.

Our GOAL-agent thus keeps two databases, respectively called the
{\em belief base} and the {\em goal base}. The difference between these
two databases originates from the different meaning assigned to sentences
stored in the belief base and sentences stored in the goal base. To clarify
the interaction between beliefs and goals, one of the more important problems
that needs to be solved is establishing a meaningful relationship between
beliefs and goals. This problem is solved here by imposing a constraint on
mental states that is derived from the default commitment strategy that agents
use. The notion of a commitment strategy is explained in more detail below.
The constraint imposed on mental states requires that an agent does not
believe that $\phi$ is the case if it has a goal to achieve $\phi$, and,
moreover, requires $\phi$ to be consistent if $\phi$ is a goal.

\begin{definition}\label{def:mentalstate} {\em (mental state)}\\ 
A {\em mental state} of an agent is a pair $\langle\bbase,\gbase\rangle$
where $\bbase\subseteq\La$ are the agent's beliefs and $\gbase\subseteq\La$
are the agent's goals (both sets may be infinite)
and $\bbase$ and $\gbase$ are such that:
\begin{itemize}
\item $\bbase$ is consistent ($\bbase\not\cmodels\false$)
\item $\gbase$ is such that, for any $\gamma \in \gbase$:
\begin{enumerate}
\item[($i)$] $\gamma$ is not entailed by the agent's beliefs ($\bbase\not\cmodels\gamma$),
\item[$(ii)$] $\gamma$ is consistent ($\not\cmodels\neg\gamma$), and
\item[$(iii)$]
for any $\gamma'$, if $\models_c \gamma \rightarrow \gamma'$ and 
$\gamma'$ satisfies $(i)$ and $(ii)$ above, then $\gamma'\in\gbase$
\end{enumerate}
\end{itemize}
\end{definition}

A mental state does {\em not} contain a program or plan component in the
`classical' sense. Although both the beliefs and the goals of an agent are
drawn from the same logical language, as we will see below, the formal
meaning of beliefs and goals is very different. This difference in meaning
reflects the different features of the beliefs and the goals of an agent.
The declarative goals are best thought of as {\em achievement} goals in this
paper. That is, these goals {\em describe} a goal state that the agent desires
to reach. Mainly due to the temporal features of such goals many properties
of beliefs fail for goals. For example, the fact that an agent has the
goal to {\em be at home} and the goal to {\em be at the movies} does not allow
the conclusion that this agent also has the conjunctive goal to {\em be at
home and at the movies} at the same time. As a consequence, less stringent
consistency requirements are imposed on goals than on beliefs. An agent may
have the goal to be at home and the goal to be at the movies simultaneously;
assuming these two goals cannot consistently be achieved at the same time
does not mean that an agent cannot have adopted both in the language GOAL.

In this paper, we assume that the language $\La$ used for representing beliefs
and goals is a simple {\em propositional language}. As a consequence, we do
not discuss the use of variables nor parameter mechanisms. Our motivation for
this assumption is the fact that we want to present our main ideas in their
simplest form and do not want to clutter the definitions below with details.
Also, more research is needed to extend the programming language with a
parameter passing mechanism, and to extend the programming logic for GOAL
with first order features.

The language $\La$ for representing beliefs and goals is extended to
a new language $\La_M$ which enables us to formulate conditions on the mental
state of an agent. The language $\La_M$ consists of so called {\em mental
state formulas}. A mental state formula is a boolean combination of the basic
mental state formulas $\bel\phi$, which expresses that $\phi$ is believed
to be the case, and $\goal\phi$, which expresses that $\phi$ is a goal of
the agent.

\begin{definition} {\em (mental state formula)}\\
The set of {\em mental state formulas} $\La_M$ is defined by:
\begin{itemize}
\item if $\phi\in\La$, then $\bel\phi\in \La_M$,
\item if $\phi\in\La$, then $\goal\phi\in \La_M$,
\item if $\varphi_1,\varphi_2\in \La_M$, then
	$\neg\varphi_1,\varphi_1\wedge\varphi_2\in \La_M$.
\end{itemize}
\end{definition}

The usual abbreviations for the propositional operators $\vee$, $\impl$, and
$\bimpl$ are used. We write $\true$ as an abbreviation for $\bel(p\vee\neg p)$
for some $p$ and $\false$ for $\neg\true$.

The semantics of belief conditions $\bel\phi$, goal conditions $\goal\phi$
and mental state formulas is defined in terms of the classical 
consequence relation
$\cmodels$.

\begin{definition}\label{def:mentalsem} {\em (semantics of mental state formulas)}\\
Let $\langle\bbase,\gbase\rangle$ be a mental state.
\begin{itemize}
\item $\langle\bbase,\gbase\rangle\mmodels\bel\phi$ iff $\bbase\cmodels\phi$,
\item $\langle\bbase,\gbase\rangle\mmodels\goal\psi$ iff
	$\psi \in \gbase$,
\item $\langle\bbase,\gbase\rangle\mmodels\neg\varphi$ iff
	$\langle\bbase,\gbase\rangle\not\mmodels\varphi$,
\item $\langle\bbase,\gbase\rangle\mmodels\varphi_1\wedge\varphi_2$ iff
	$\langle\bbase,\gbase\rangle\mmodels\varphi_1$ and
	$\langle\bbase,\gbase\rangle\mmodels\varphi_2$.
\end{itemize}
\end{definition}

We write $\mmodels \varphi$ for the fact that mental state formula
$\varphi$ is true in all mental states $\langle\bbase,\gbase\rangle$.

A number of properties of the belief and goal modalities and the relation
between these operators are listed in Tables~\ref{table:beliefs}
and \ref{table:goals}. Here, $\cvdash$ denotes derivability in 
classical logic, whereas $\mvdash$ refers to derivability in the
language of mental state formulas ${\cal L}_M$.

The first rule ($R1$) below states that mental state formulas 
that `have the form of a classical tautology'
(like $(\bel\phi \lor \neg\bel\phi)$ and $\goal \phi_1
\rightarrow (\bel\phi_2 \rightarrow \goal\phi_1)$), are also 
derivable in $\mvdash$.
By the necessitation
rule ($R2$), an agent believes all classical tautologies. 
Then, ($A1$) expresses that the belief
modality distributes over implication. This implies that the beliefs of 
an agent are closed under logical
consequence. Finally, $A2$ 
states that the beliefs of an agent are consistent. 
In essence, the belief operator thus satisfies the
properties of the system {\bf KD} (see \cite{fahamova94,MeyHoe94a}).
Although in its current presentation, our language does not allow
for nested (belief-) operators, 
from \cite[Section 1.7]{MeyHoe94a} we conclude that we may assume
{\em as if} our agent has positive ($\bel\phi \rightarrow \bel\bel\phi$)
and negative ($\neg\bel\phi \rightarrow \bel\neg\bel\phi$) introspective
properties: every formula in the system {\bf KD45} (which is {\bf KD} 
together with the two mentioned properties) is equivalent to a formula
without nestings of operators.

\begin{table}[!h]
\begin{center}
\fbox{\begin{minipage}{10.5cm}
\begin{tabular}{ll}
 $R1$ & if $\varphi$ is an instantiation of a classical tautology, then $\mvdash \varphi$\\
 $R2$ & $\cvdash\phi\Impl\mvdash\bel\phi$, for $\phi\in\La$\\
 $A1$ & $\mvdash\bel(\phi\impl\psi)\impl(\bel\phi\impl\bel\psi)$\\
 $A2$ & $\mvdash\neg\bel\false$\\
\end{tabular}
\end{minipage}}
\end{center}
\caption{Properties of Beliefs}
\label{table:beliefs}
\end{table}
Axiom $A4$ below, is a consequence of the constraint on mental
states and expresses that if an agent believes $\phi$ it does not have a goal
to achieve $\phi$. 
As a consequence, an agent cannot have a goal to achieve a
tautology: $\neg\goal\true$. 
An agent also does not have inconsistent goals ($A3$), that is,
$\neg\goal\false$ is an axiom (see Table~\ref{table:goals}).
Finally, the conditions that allow to conclude
that the agent has a (sub)goal $\psi$ are that the agent has a goal
$\phi$ that logically entails $\psi$ and that the agent does not believe
that $\psi$ is the case. Axiom $A5$ below then allows to conclude
that $\goal\psi$ holds. From now on, for any mental state formula $\varphi$,
$\mvdash \varphi$ means that that there is a derivation of $\varphi$ using the
proof rules $R1$ and $R2$ and the axioms $A1 -- A6$. If $\Delta$ is a set
of mental state formulas from $\La_M$, then $\Delta \mvdash \varphi$ means
that there is a derivation of $\varphi$ using the rules and axioms mentioned,
and the formulas of $\Delta$ as premises.

\begin{table}[h]
\begin{center}
\fbox{\begin{minipage}{10.5cm}
\begin{tabular}{ll}
$A3$ & $\mvdash\neg\goal\false$\\
$A4$ & $\mvdash\bel\phi\impl\neg\goal\phi$\\ 
$A5$ & $\cvdash\phi\impl\psi \Rightarrow \ \mvdash
\neg\bel\psi\impl(\goal\phi\impl\goal\psi)$\\
\end{tabular}
\end{minipage}}
\end{center}
\caption{Properties of Goals}
\label{table:goals}
\end{table}
The goal modality is a weak logical operator. For example, the goal
modality does not distribute over implication. A counter example
is provided by the goal base that is generated from $\{p, p\impl q\}$. 
The consequences of goals are only computed {\em locally}, from 
individual goals. But even from the goal base $\{p \land (p \impl q)$
one cannot conclude that $q$ is a goal, since this conclusion is 
blocked in a mental state in which $q$ is already believed.
Deriving only consequences of goals locally
ensures that from the fact that $\goal\phi$ and
$\goal\psi$ hold, it is not possible to conclude that
$\goal(\phi\wedge\psi)$. This reflects the fact that individual goals
cannot be added to a single bigger goal; recall that two individual goals
may be inconsistent ($\goal\phi\wedge\goal\neg\phi$ is satisfiable) in
which case taking the conjunction would lead to an inconsistent goal.
In sum, most of the usual problems that many logical operators for
motivational attitudes suffer from do not apply to our $\goal$ operator
(cf.\ also \cite{Mey99}). On the other hand, the last property of
Lemma~\ref{lem:nongoals} justifies to call $\goal$ a logical, and 
not just a syntactical operator:

{\begin{lemma}\strut
\label{lem:nongoals}
\begin{itemize}
\item $\not\mmodels\goal(\phi\impl\psi)\impl(\goal\phi\impl\goal\psi)$,
\item $\not\mmodels\goal(\phi\wedge (\phi\impl\psi))\impl\goal\psi$,
\item $\not\mmodels(\goal\phi\wedge\goal\psi)\impl\goal(\phi\wedge\psi)$
\item $\cmodels (\varphi \leftrightarrow \psi) \ \Rightarrow\ 
       \mmodels (\goal\varphi \leftrightarrow \goal\psi)$
\end{itemize}
\end{lemma}
}

One finds a similar quest for such weak 
operators in {\em awareness logics\/} for doxastic and epistemic
modalities, see e.g. \cite{faghal88,Thijsse93}. As agents do not
want all the side-effects of their goals, being limited reasoners
they also do not always adopt all the logical consequences of their
belief or knowledge. However, the question remains whether
modal logic is {\em the\/} formal tool to reason with and about
goals. Allowing explicitly for mutually inconsistent goals, our
treatment of goals resides in the landscape of
{\em paraconsistent logic} (cf.~\cite{gries89}).
One might even go a step further and explore to use {\em linear logic\/} (\cite{girard87} to reason about goals, enabling to have
the same goal more than once, and to model process and resource 
use in a fine-tuned way. We will not pursue the different options
for logics of goals in this paper.

\begin{theorem}\label{thm:soundmvdash}
({\em Soundness and Completeness of $\mvdash$})\\
For any $\varphi \in \La_M$, we have
\[ \mvdash \varphi \Leftrightarrow \mmodels \varphi \]
\end{theorem}
{\bf Proof}. We leave it for the reader to check soundness (i.e., the
`$\Rightarrow$'-direction). Here, $A1$ and $A2$ are immediate consequences
of the definition of a belief as a consequence from a given consistent set, 
$A3$ follows from condition $(ii)$ of Definition \ref{def:mentalstate},
$A4$ from property $(i)$ and $A5$ from $(iii)$ of that same definition.\\
For completeness, assume that $\not\mvdash\varphi$. Then $\neg\varphi$ is
consistent, and we will construct a mental state $\mstate$
that verifies $\varphi$. First, we build a maximal 
$\mvdash$-consistent set $\Delta$
with $\neg\varphi \in \Delta$. This $\Delta$ can 
be split in a set $\bbase$
and a set $\gbase$ as follows: 
$\bbase = \{\phi | \bel\phi\in\Delta\}$
and $\gbase = \{\phi | \goal\phi\in\Delta\}$. 
We now prove two properties of $\mstate$:
\begin{enumerate}
\item
$\mstate$ is a mental state
\item
$\mstate$ satisfies the following coincidence property:
\[ \mbox{for all } \chi\in\La_m:\ \mstate \mmodels \chi \Leftrightarrow 
\chi \in \Delta \]
\end{enumerate}
The proofs for these claims are as follows:
\begin{enumerate}
\item
We must show that $\mstate$ satisfies the properties of 
Definition~\ref{def:mentalstate}. Obviously, $\bbase$ is classically
consistent, since otherwise we would have $\bel\bot$ in the $\mvdash$-consistent
set $\Delta$, which is prohibited by axiom $A2$.
Also, by axiom $A3$, no $\gamma\in \gbase$ is equivalent to $\bot$.
We now show that no $\gamma\in \gbase$ is classically entailed by $\bbase$.
Suppose that we would have that $\sigma_1, \dots, \sigma_n
\cvdash \gamma$, for certain $\sigma_1, \dots, \sigma_n \in \bbase$
and $\goal\in\gbase$.
Then, by construction of $\bbase$ and $\gbase$, 
the formulas $\bel\varphi_1, \dots, \bel\varphi_n, \goal\gamma$
all are members of the maximal $\mvdash$-consistent set $\Delta$.
Since $\sigma_1, \dots, \sigma_n
\cvdash \gamma$, by the deduction theorem for $\cvdash$, 
$R2$ and $A2$ we conclude 
$\mvdash(\bel\sigma_1, \dots, \bel\sigma_n) \rightarrow
\bel\gamma$. But this means that both $\bel\gamma$ and $\goal\gamma$
are members of $\Delta$, which is prohibited by axiom $A4$.
Finally, we show $(iii)$ of Definition~\ref{def:mentalstate},
Suppose $\gamma\in \gbase$, $\cmodels\gamma \rightarrow \gamma'$
and that $\gamma`$ is consistent, and not classically entailed by $\bbase$.
We have to $\gamma' \in \gbase$, and this is immediately guaranteed
by axiom $A5$.

\item
The base case for the second claim 
is about $\bel\phi$ and $\goal\phi$, with
$\phi \in \La$. We have $\mstate \mmodels \bel\phi$ iff 
$\bbase \cmodels \phi$ iff, by definition of $\bbase$, $\{\sigma | 
\bel\sigma \in \Delta\} \cmodels \phi$. Using compactness and 
the deduction theorem for classical logic, we find 
$\cvdash \sigma_1 \land \cdots \land \sigma_n) \rightarrow \phi$, 
for some propositional formulas $\sigma_1, \dots, \sigma_n$.
By the rule $R2$ we conclude 
$\mvdash \bel(\sigma_1 \land \cdots \land \sigma_n) \rightarrow \phi)$.
By $A1$, this is equivalent to 
$\mvdash (\bel\sigma_1 \land \cdots \land \bel\sigma_n) \rightarrow \bel\phi$
and, since all the $\bel\sigma_i (i \leq n)$ are members of $\Delta$,
we have $\bel\phi \in \Delta$.
For the other base case, consider $\mstate \mmodels \goal\phi$,
which, using the truth-definition for $\goal$, holds iff
$\gamma \in \Gamma$. By definition of $\Gamma$, this means
that $\goal\gamma \in \Delta$, which was to be proven.
The cases for negation and conjunction follow immediately 
from this.
Hence, in particular, we have $\mstate \mmodels \neg\varphi$, and thus
$\not\mmodels \varphi$.
\end{enumerate}

\subsection{{GOAL} Agents}
A third basic concept in GOAL is that of an agent {\em capability}. The
capabilities of an agent consist of a set of so called {\em basic actions}.
The effects of executing such a basic action are reflected in the beliefs of
the agent and therefore a basic action is taken to be a belief update on
the agent's beliefs. A basic action thus is a {\em mental state transformer}.
Two examples of agent capabilities are the actions $\myins(\phi)$ for
inserting $\phi$ in the belief base and $\mydel(\phi)$ for removing $\phi$
from the belief base. Agent capabilities directly affect the
belief base of the agent and not its goals,
but because of the constraints on mental states they may
as a side effect modify the current goals. For the purpose of modifying the
goals of the agent, two special actions $\adopt(\phi)$ and $\drop(\phi)$ are
introduced to respectively adopt a new goal or drop some old goals. We write
$Bcap$ and use it to denote the set of all belief update capabilities of an
agent. $Bcap$ thus does not include the two special actions for goal updating
$\adopt(\phi)$ and $\drop(\phi)$. The set of all capabilities is then defined
as $Cap=Bcap\cup\{\adopt(\phi), \drop(\phi) | \phi\in\La\}$. Individual
capabilities are denoted by $\ac$.

The set of basic actions or capabilities associated with an agent determines
what an agent {\em is able to do}. It does not specify {\em when} such a
capability should be exercised and when performing a basic action is to the
agent's advantage. To specify such conditions, the notion of a {\em conditional
action} is introduced. A conditional action consists of a mental state
condition expressed by a mental state formula and a basic action. The mental
state condition of a conditional action states the conditions that must hold
for the action to be selected. Conditional actions are denoted by the symbol
$b$ throughout this paper.

\begin{definition} {\em (conditional action)}\\
A {\em conditional action} is a pair $\varphi\impl do({\sf a})$ such that
$\varphi\in \La_M$ and ${\sf a}\in Cap$.
\end{definition}

Informally, a conditional action $\varphi\impl do({\sf a})$ means that
if the mental condition $\varphi$ holds, then the agent may consider doing
basic action $\ac$. Of course, if the mental state condition holds in the
current state, the action $\ac$ can only be successfully executed if the
action is {\em enabled}, that is, only if its precondition holds.

A GOAL agent consists of a specification of an {\em initial mental state} and
a set of conditional actions.

\begin{definition} {\em (GOAL agent)}\\
A {\em GOAL agent} is a triple $\langle\Pi,\bbase_0,\gbase_0\rangle$
where $\Pi$ is a non-empty set of conditional actions, and
$\langle\bbase_0,\gbase_0\rangle$ is the initial mental state.
\end{definition}

\subsection{The Operational Semantics of GOAL}

One of the key ideas in the semantics of GOAL is to incorporate into the
semantics a particular {\em commitment strategy} (cf.\ \cite{Rao90,Coh90}).
The semantics is based on a particularly simple and transparent commitment
strategy, called {\em blind commitment}. An agent that acts according to a
blind commitment strategy drops a goal if and only if it believes that that
goal has been achieved. By incorporating this commitment strategy into the
semantics of GOAL, a {\em default} commitment strategy is built into agents.
It is, however, only a default strategy and a programmer can overwrite this
default strategy by means of the {\sf drop} action. It is not possible,
however, to adopt a goal $\phi$ in case the agent believes that $\phi$ is
already achieved.

The semantics of action execution should now be defined in conformance with
this basic commitment principle. Recall that the basic capabilities of an
agent were interpreted as belief updates. Because of the default commitment
strategy, there is a relation between beliefs and goals, however, and we
should extend the belief update associated with a capability to a mental
state transformer that updates beliefs as well as goals according to the
blind commitment strategy. To get started, we thus assume that some
specification of the belief update semantics of all capabilities - except
for the two special actions $\adopt$ and $\drop$ which only update goals -
is given. Our task is, then, to construct a mental state transformer semantics
from this specification for each action. That is, we must specify how a basic
action updates the complete current mental state of an agent starting with
a specification of the belief update associated with the capability only.

From the default blind commitment strategy, we conclude that if a basic action
$\ac$ - different from an $\adopt$ or $\drop$ action - is executed, then a goal
is dropped {\em only if} the agent {\em believes} that the goal has been
accomplished after doing $\ac$. The revision of goals thus is based on the
beliefs of the agent. The beliefs of an agent represent all the information
that is available to an agent to decide whether or not to drop or adopt a
goal. So, in case the agent believes that a goal has been achieved by
performing some action, then this goal must be removed from the current
goals of the agent. Besides the default commitment strategy, only the two
special actions $\adopt$ and $\drop$ can result in a change to the goal base.

The initial specification of the belief updates associated with the
capabilities $Bcap$ is formally represented by a partial function $\Trans$
of type $:Bcap\times\wp(\La)\arr\wp(\La)$. $\Trans(\ac,\bbase)$ returns the
result of updating belief base $\bbase$ by performing action $\ac$. The fact
that $\Trans$ is a {\em partial} function represents the fact that an action
may not be {\em enabled} or executable in some belief states. The mental
state transformer function $\update$ is derived from the semantic function
$\Trans$ and also is a partial function. As explained,
$\update(\ac,\mstate)$ removes
any goals from the goal base $\gbase$ that have been achieved by doing $\ac$.
The function $\update$ also defines the semantics of the two special actions
$\adopt$ and $\drop$. An $\adopt(\phi)$ action adds $\phi$ to the goal base
if $\phi$ is consistent and $\phi$ is not believed to be the case. A
$\drop(\phi)$ action removes every goal that entails $\phi$ from the goal
base. As an example, consider the two extreme cases: $\drop(\false)$
removes no goals, whereas $\drop(\true)$ removes all current goals.

\begin{definition} {\em (mental state transformer $\update$)}\\
Let $\langle\bbase,\gbase\rangle$ be a mental state, and $\Trans$ be
a partial function that associates belief updates with agent capabilities.
Then the partial function $\update$ is defined by:
\begin{tabular}{lll}
$\update(\ac,\langle\bbase,\gbase\rangle)$ & $=$ &
	        $\langle\Trans(\ac,\bbase),\gbase\setminus
	\{\psi\in\gbase\;|\;\Trans(\ac,\bbase)\cmodels\psi\}\rangle$\\
& & 	for $\ac\in Bcap$, if $\Trans(\ac,\bbase)$ is defined\\
$\update(\ac,\langle\bbase,\gbase\rangle)$ & &
      is undefined for $\ac\in Bcap$ if $\Trans(\ac,\bbase)$
	is undefined\\
$\update(\drop(\phi),\langle\bbase,\gbase\rangle)$ & $=$ &
	$\langle\bbase,\gbase\setminus
	\{\psi\in\gbase\;|\;\psi\cmodels\phi\}\rangle$\\
$ \update(\adopt(\phi),\langle\bbase,\gbase\rangle) $ & $=$&
$	\langle\bbase,\gbase\cup\{\phi' | \Sigma\not\mmodels \phi'\
 \&\ \cmodels \phi \rightarrow \phi' \}\rangle$\\
& & 	if $\not\cmodels\neg\phi$ and $\Sigma\not\cmodels\phi$\\
$\update(\adopt(\phi),\langle\bbase,\gbase\rangle)$ & &
  is undefined if $\bbase\cmodels\phi$ or $\cmodels\neg\phi$\\
\end{tabular}
\end{definition}

The semantic function $\update$ maps an agent capability and a mental state
to a new mental state. The capabilities of an agent are thus interpreted as
{\em mental state transformers} by $\update$. Although it is not allowed to
adopt a goal $\phi$ that is inconsistent - an $\adopt(\false)$ is not enabled
- there is no check on the global consistency of the goal base of an agent
built into the semantics. This means that it is allowed to adopt a new goal
which is inconsistent with another goal present in the goal base. For example,
if the current goal base $\Gamma$ contains $p$, it is legal to execute
the action $\adopt(\neg p)$ resulting in a new goal base containing
$p, \neg p$, (if $\neg p$ was not already believed).
Although inconsistent goals cannot be achieved at the same time, they may be
achieved in some temporal order. Individual goals in the goal base, however,
are required to be consistent. Thus, whereas local consistency is required
(i.e. individual goals must be consistent), global consistency of the goal
base is not required.

The second idea incorporated into the semantics concerns the {\em selection
of conditional actions}. A conditional action $\varphi\impl do({\sf a})$
may specify conditions on the beliefs as well as conditions on the goals of
an agent. As is usual, conditions on the beliefs are taken as a precondition
for action execution: only if the agent's current beliefs entail the belief
conditions associated with $\varphi$ the agent will select ${\sf a}$ for
execution. The goal condition, however, is used in a different way. It is
used as a means for the agent to determine whether or not the action will
help bring about a particular goal of the agent. In short, the goal
condition specifies where the action is good for. This 
does not mean that the action necessarily establishes the goal immediately,
but rather may be taken as an indication that the action is helpful in
bringing about a particular state of affairs. 

\weg{
\begin{definition} {\em ($\phi$ partially fulfils a goal in a mental state)}\\
Let $\langle\bbase,\gbase\rangle$ be a mental state, and $\phi\in\La$.
Then:
\[
\phi\leadsto_\bbase\gbase \mbox{ iff for some }\psi\in\gbase:
	\psi\cmodels\phi
	\mbox{ and } \bbase\not\cmodels\phi
\]
\end{definition}

Informally, the definition of $\phi\leadsto_\bbase\gbase$ can be paraphrased
as follows: the agent needs to establish $\phi$ to realise one of its goals
in $\gamma$, but does not believe that $\phi$ is the case.  The formal
definition of $\phi\leadsto_\bbase\gamma$ entails that the realisation of
$\phi$ would bring about at least part of one of the goals in the goal
base $\gbase$ of the agent. The condition that $\phi$ is not entailed by
the beliefs of the agent ensures that a goal is not a tautology. Of course,
variations on this definition of the semantics of goals are conceivable. For
example, one could propose a stronger definition of $\leadsto$ such that $\phi$
brings about the {\em complete} realisation of a goal in the current goal base
$\gbase$ instead of just part of such a goal. However, our definition of
$\leadsto$ provides for a simple and clear principle for action selection:
the action in a conditional action is only executed in case the goal condition
associated with that action partially fulfils some goal in the current goal
base of the agent.
}

\weg{
Now we have defined the formal semantics of mental state formulas, we are
able to formally define the selection and execution of a conditional action.
The selection of an action by an agent depends on the satisfaction conditions
of the mental state condition associated with the action in a conditional
action. The conditions for action selection thus may express conditions on
both the belief and goal base of the agent. The belief conditions associated
with the action formulate preconditions on the current belief base of the
agent. Only if the current beliefs of the agent satisfy these conditions, an
action may be selected. A condition $\goal\phi$ on the goal base is satisfied
if $\phi$ is entailed by one of the current goals of the agent
(and thus, assuming the programmer did a good job, helps in bringing about
one of these goals). The intuition here is that an agent is satisfied with
anything bringing about at least (part of) one of its current goals. Note
that a condition $\goal\phi$ can only be satisfied if the agent does not
already believe that $\phi$ is the case ($\bbase\not\cmodels\phi$) which
prevents an agent from performing an action without any need to do so.
}

In the definition below, we assume that the action component $\Pi$ of an
agent $\langle\Pi,\Sigma_0,\Gamma_0\rangle$ is fixed. The execution of an
action gives rise to a {\em computation step} formally denoted by the
transition relation $\stackrel{b}{\lra}$ where $b$ is the conditional
action executed in the computation step. More than one computation step
may be possible in a current state and the step relation $\lra$ thus
denotes a {\em possible} computation step in a state. A computation step
updates the current state and yields the next state of the computation.
Note that because $\update$ is a partial function, a conditional action
can only be successfully executed if both the condition is satisfied and
the basic action is enabled.

\begin{definition} {\em (action selection)}\\
Let $\mstate$ be a mental state and
$b=\varphi\impl do(\ac)\in\Pi$. Then, as a rule, we have:\\
If
\begin{itemize}
\item the mental condition $\varphi$ holds in $\mstate$,
	i.e. $\mstate\models\varphi$, and
\item $\ac$ is enabled in $\mstate$, i.e.
	$\update({\sf a},\langle\bbase,\gbase\rangle)$ is defined,
\end{itemize}
then
$\langle\bbase,\gbase\rangle\stackrel{b}{\lra}
  \update({\sf a},\langle\bbase,\gbase\rangle)$ is a possible computation
step. The relation $\lra$ is the smallest relation closed under this rule.
\end{definition}

Now, the semantics of GOAL agents is derived directly from the operational
semantics and the computation step relation $\lra$ . 
The meaning of a GOAL agent consists of a set of
so called {\em traces}. A trace is an infinite computation sequence of
consecutive mental states interleaved with the actions that are scheduled for
execution in each of those mental states. The fact that a conditional action
is scheduled for execution in a trace does not mean that it is also enabled
in the particular state for which it has been scheduled. In case an action
is scheduled but not enabled, the action is simply skipped and the resulting
state is the same as the state before. In other words, enabledness
is not a criterion for selection, but rather it decides whether
something is happening in a state, once selected.

\begin{definition} {\em (trace)}\\
A trace $s$ is an infinite sequence $s_0,b_0,s_1,b_1,s_2,\ldots$ such that
$s_i$ is a mental state, $b_i$ is a conditional action, and for every $i$
we have: $s_i\stackrel{b_i}{\lra}s_{i+1}$, or $b_i$ is not enabled in
$s_i$ and $s_i=s_{i+1}$.
\end{definition}

An important assumption in the semantics for GOAL is a {\em fairness}
assumption. Fairness assumptions concern the fair selection of actions during
the execution of a program. In our case, we make a {\em weak fairness}
assumption \cite{Man92}. A trace is weakly fair if it is not the case that an
action is always enabled from some point in time on but is never selected for
execution. This weak fairness assumption is built into the semantics by
imposing a constraint on traces. By definition, a {\em fair trace} is a
trace in which each of the actions is scheduled infinitely often. In a fair
trace, there always will be a future time point at which an action is
scheduled (considered for execution) and by this scheduling policy a fair
trace implements the weak fairness assumption. However, note that the fact
that an action is scheduled does not mean that the action also is enabled
(and therefore, the selection of the action may result in an idle step which
does not change the state).

The meaning of a GOAL agent now is defined as the set of fair traces in which
the initial state is the initial mental state of the agent and each of the
steps in the trace corresponds to the execution of a conditional action or
an idle transition.

\begin{definition} {\em (meaning of a GOAL agent)}\\
The meaning of a GOAL agent $\langle \Pi, \bbase_0,\gbase_0\rangle$
is the set of {\em fair} traces $S$ such that for $s\in S$ we have
$s_0=\langle\bbase_0,\gbase_0\rangle$.
\weg{
\item and it is not the case that an action $b\in\Pi$ is infinitely
	often enabled in $s$ and for all $i\geq n$
	$s_i\not\stackrel{b}{\lra}s_{i+1}$.
\end{itemize}}
\end{definition}

\subsection{Mental States and Enabledness}
We formally said that a capability $\ac\in Cap$ is {\em enabled} in a mental state
$\mstate$ in case
$\update(\ac,\mstate)$ is defined. This definition
implies that a
belief update capability $\ac\in Bcap$ is enabled if $\Trans(\ac,\Sigma)$
is defined. 
Let us assume that this only depends on the action $\ac$ --this
seems reasonable, since a paradigm like AGM (\cite{agm}) only requires
that a revision with $\varphi$ fails iff $\varphi$ is classically
inconsistent, whereas expansions and contractions succeed for
all $\varphi$, hence the question whether such an operation is
enabled does not depend on the current beliefs.
A conditional action $b$ is {\em enabled} in a mental state
$\langle\bbase,\gbase\rangle$ if there are $\bbase',\gbase'$ such that
$\langle\bbase,\gbase\rangle\stackrel{b}{\lra}\langle\bbase',\gbase'\rangle$.
Note that if a capability $\ac$ is not enabled, a conditional action
$\varphi\impl do(\ac)$ is also not enabled. The special predicate $enabled$
is introduced to denote that a capability $\ac$ or conditional action $b$
is enabled (denoted by $enabled(\ac)$ respectively $enabled(b)$).

The relation between the enabledness of capabilities and conditional actions
is stated in the next table together with the fact that $\drop(\phi)$ is always
enabled and a proof rule for deriving $enabled(\adopt(\phi))$. 
Let $\La_{ME}$ be the language obtained by Boolean combinations
of mental state formulas and enabledness formulas.
We denote
derivability in the system for this language by $\vdash_{ME}$.
Then, $\vdash_{ME}$ consists of the axioms and rules for $\mvdash$,
plus

\begin{table}[!h]
\begin{center}
\fbox{\begin{minipage}{10.5cm}
\begin{tabular}{ll}
$E1$ & $ enabled(\varphi\impl do(\ac))\bimpl(\varphi\wedge enabled(\ac))$,\\
$E2$ & $ enabled(\drop(\phi))$,\\
$R3$ & $\not\cmodels\neg\phi \Rightarrow \vdash_{ME}
	\neg\bel\phi\leftrightarrow enabled(\adopt(\phi)$\\
$R4$ &
     $\cmodels \neg\phi \Rightarrow \vdash_{ME} \neg\enabled(\adopt(\phi))$\\
$R5$ & $\vdash_{ME} enabled(\ac)$ if $\Trans(\ac,\cdot)$ is defined ($\ac \in Bcap$)
\end{tabular}
\end{minipage}}
\end{center}
\caption{Enabledness}
\label{table:enabledness}
\end{table}

Rule $R5$ enforces that we better write $\vdash^\Trans_{ME}$ given
a belief revision function $\Trans$, but in the sequel
we will suppress this $\Trans$.
The semantics $\models_{ME}$ for $\La_{ME}$ is based on 
truth in pairs $\mstate, \Trans$, where $\mstate$ is a mental state
and $\Trans$ a partial function for belief updates. For formulas
of the format $\bel \varphi$ and $\goal\varphi$, we just use the
mental state and Definition~\ref{def:mentalsem} to determine their
truth. For enabledness formulas, we have the following:

\begin{definition}(Truth of enabledness)\label{def:enabledsem}
\begin{itemize}
\item
$\mstate, \Trans \models_{ME} \enabled{(\ac)}$ iff 
$\Trans(\ac,\bbase)$ is defined
\item
$\mstate, \Trans \models_{ME} \enabled(\drop(\phi))$ iff true
\item
$\mstate, \Trans \models_{ME} \enabled(\adopt(\phi))$ iff 
$\not\cmodels \neg\phi$ and $\mstate, \Trans \models_{ME} \neg\bel\phi$
\item
$\mstate, \Trans \models_{ME} \enabled({\varphi \rightarrow do(\ac)})$ iff 
$\mstate, \Trans \models_{ME} \phi$ and at the same time $\mstate, \Trans \models_{ME} \enabled(\ac)$
\end{itemize}

Note that we can summarize this definition to:
\begin{itemize}
\item $\mstate, \Trans \models_{ME} enabled(\ac)$ iff
	$\update(\ac,\mstate)$ is defined for $\ac\in Cap$,
\item $\mstate,\Trans\models_{ME} enabled(b)$ iff
	$\models_{ME} \varphi$ and 
	there are $\Sigma',\Gamma'$ such that
	$\langle\bbase,\gbase\rangle\stackrel{b}{\lra}
	\langle\bbase',\gbase'\rangle$ for conditional actions where
	$b=\varphi\impl do(\ac)$.
\end{itemize}
\end{definition}

\begin{theorem}(Soundness and Completeness of $\vdash_{ME}$)\\
We have, for all formulas $\varphi$ in $\La_{ME}$,
\[
\vdash_{ME} \varphi \mbox{ iff } \models_{ME} \varphi
\]
\end{theorem}
{\bf Proof}. Again, checking soundness is straightforward and left to the reader.
For the converse, we have to make a complexity measure
explicit for $\La_{ME}$-formulas, along which the induction can proceed.
It suffices to stipulate that the complexity of 
$\enabled(\psi \rightarrow do(\ac)$ is greater than that
of $\bel\psi$ and $\enabled(\ac)$. Furthermore, the complexity
of $\enabled(\adopt(\phi)$ is greater than that of 
$(\neg)\bel\phi$.
Now, suppose that $\not\vdash_{ME} \varphi$, i.e., 
$\neg \varphi$ is consistent. Note that the language $\La_{ME}$
is countable, so that we can by enumeration, extend $\{ \neg\varphi\}$
to a maximal $\vdash_{ME}$-consistent set $\Delta$. From this
$\Delta$, we distill a pair $\mstate, \Trans$ as follows:
$\bbase = \{\varphi | \bel\varphi \in \Delta\}$, $\gbase = \{\varphi |
\goal\varphi \in \Delta\}$, and $\Trans(\ac,\bbase)$ is defined
iff $\enabled(\ac) \in \Delta$, for any belief capability $\ac$.
We claim, for all $\chi \in \La_{ME}$:
\[
\mstate, \Trans \models \chi \mbox{ iff } \chi \in \Delta
\]

For formulas of type $\bel\psi$ and $\goal\psi$ this is easily seen.
Let us check it for enabledness formulas. 
\begin{itemize}
\item
$\chi = \enabled(\ac)$, with $\ac$ a belief capability. 
By construction of $\Trans$, the result immediately holds
\item
$\chi = \enabled(\drop(\phi))$. By construction of $\Delta$,
every $\enabled(\drop(\phi))$ is an element of $\Delta$ (because
of axiom $E2$), and also, every such formula is true in $\mstate,\Trans$.
\item
$\chi = \enabled(\adopt(\phi))$. Suppose $\mstate,\Trans \models_{ME}
\chi$. Then, $\not\models \neg\phi$ and $\mstate,\Trans\not\models_{ME}\bel\phi$.
By the induction hypothesis, we have that $\bel\phi \not\in \Delta$,
hence $\neg\bel\phi \in \Delta$, and, by $R3$, $\enabled(\adopt\phi)) \in \Delta$.
For the converse, suppose $\enabled(\adopt(\phi)) \in \Delta$. 
Then (by $R4$), we cannot have that $\cmodels \neg\phi$.
Hence, $\not\cmodels\neg\phi$, and by $R3$, we also have $\neg\bel\phi
\in \Delta$ and hence, by applying the 
induction hypothesis, $\mstate,\Trans\cmodels\neg\bel\phi$. Since
$R3$ is a sound rule, we finally
conclude that $\mstate,\Trans\models_{ME} \enabled(\adopt(\phi))$.
\item
$\chi = \enabled(\psi \rightarrow \ac)$. We
can write this as $\psi \land \enabled(\ac)$ and then use the induction
hypothesis.
\end{itemize}

\section{A Personal Assistant Example}
In this section, we give an example to show how the programming language GOAL
can be used to program agents. The example concerns a shopping agent that
is able to buy books on the Internet on behalf of the user. The example
provides for a simple illustration of how the programming language works.
The agent in our example uses a standard procedure for buying a book. It
first goes to a bookstore, in our case Am.com. At the web site of
Am.com it searches for a particular book, and if the relevant page with
the book details shows up, the agent puts the book in its shopping cart. In
case the shopping cart of the agent contains some items, it is allowed to buy
the items on behalf of the user. The idea is that the agent adopts a goal
to buy a book if the user instructs it to do so.

The set of capabilities $Bcap$ of the agent is defined by
\[\{ goto\_website(site), search(book), put\_in\_shopping\_cart(book),
	pay\_cart \}
\]
The capability $goto\_website(site)$ goes to the selected web page $site$.
In our example, relevant web pages are the home page of the user, the main
page of Am.com, web pages with information about books to buy, and a
web page that shows the current items in the shopping cart of the agent.
The capability $search(book)$ is an action that can be selected at the
main page of Am.com and selects the web page with information about
$book$. The action $put\_in\_shopping\_cart(book)$ can be selected
on the page concerning $book$ and puts $book$ in the cart; a new web page
called $ContentCart$ shows up showing the content of the cart. Finally,
in case the cart is not empty the action $pay\_cart$ can be selected to
pay for the books in the cart.

In the program text below, we assume that $book$ is a variable referring to
the specifics of the book the user wants to buy (in the example, we use
variables as a means for abbreviation; variables should be thought of as
being instantiated with the relevant arguments in such a way that predicates
with variables reduce to propositions). The initial beliefs of the agent are
that the current web page is the home page of the user, and that it is not
possible to be on two different web pages at the same time. We also assume that
the user has provided the agent with the goals to buy {\em The Intentional
Stance} by Daniel Dennett and {\em Intentions, Plans, and Practical Reason}
by Michael Bratman.

\begin{center}
\mbox{
\begin{minipage}{11.5cm}
\vspace{-.4cm}
\[\hspace{-1cm}
\Pi=\{\\
\bel(current\_website(hpage(user))\vee current\_website(ContentCart))\\
\t2	\wedge\goal(bought(book))\impl do(goto\_website(Am.com)),\\
\bel(current\_website(Am.com))\wedge\neg\bel(in\_cart(book))\wedge\\
\t2	\goal(bought(book))\impl do(search(book)),\\
\bel(current\_website(book))\wedge\goal(bought(book))\\
\t2 \impl
	do(put\_in\_shopping\_cart(book)),\\
\bel(in\_cart(book))\wedge \goal(bought(book))\impl do(pay\_cart)
\},\\
\hspace{-1cm}\bbase_0=\{current\_webpage(hpage(user)),\\
\t2	\forall s,s'((s\neq s'\wedge current\_webpage(s))\impl\neg
	current\_webpage(s'))\},\\
\hspace{-1cm}\gbase_0=\{bought(\mbox{The Intentional Stance})\\
\t2
\wedge
bought(\mbox{Intentions, Plans and Practical Reason})\}
\]
\end{minipage}
}
\vspace{.1cm} \\
{\em GOAL Shopping Agent}
\end{center}

Some of the details of this program will be discussed in the sequel, when
we prove some properties of the program. The agent basically follows the
recipe for buying a book outlined above. For now, however, just note
that the program is quite flexible, even though the agent more or less
executes a fixed recipe for buying a book. The flexibility results from the
agent's knowledge state and the non-determinism of the program. In particular,
the ordering in which the actions are performed by the agent - which book
to find first, buy a book one at a time or both in the same shopping cart,
etc. is not determined by the program. The scheduling of these actions thus
is not fixed by the program, and might be fixed arbitrarily on a particular 
agent architecture used to run the program.

\section{Logic for GOAL}
On top of the language GOAL and its semantics, we now construct a temporal
logic to prove properties of GOAL agents. The logic is similar to other
temporal logics but its semantics is derived from the operational semantics
for GOAL. Moreover, the logic incorporates the belief and goal modalities
used in GOAL agents. We first informally discuss  the use of Hoare
triples for the specification of actions. In Section 
\ref{subsec:complete} we give a sound an complete 
system for such triples. These Hoare triples play an
important role in the programming logic since it can be shown that
temporal properties of agents can be proven by means of proving Hoare
triples for actions only. Finally,
in \ref{subsec:temporal} the language for expressing temporal
properties and its semantics is defined and the fact that certain classes
of interesting temporal properties can be reduced to properties of actions,
expressed by Hoare triples, is proven.

\subsection{Hoare Triples}

The specification of basic actions provides the basis for the programming
logic, and, as we will show below, is all we need to prove properties of
agents. Because they play such an important role in the proof theory of GOAL,
the specification of the basic agent capabilities requires special care. In
the proof theory of GOAL, Hoare triples of the form $\{\varphi\}\;b\;\{\psi\}$,
where $\varphi$ and $\psi$ are {\em mental state formulas}, are used to specify
actions. The use of Hoare triples in a
formal treatment of traditional assignments is well-understood \cite{And91}.
Because the agent capabilities of GOAL agents are quite different from
assignment actions, however, the traditional predicate transformer semantics
is not applicable. GOAL agent capabilities are mental state transformers and,
therefore, we require more extensive basic action theories to formally
capture the effects of such actions. Hoare triples are used to specify the
{\em postconditions} and the {\em frame conditions} of actions. The
postconditions of an action specify the effects of an action whereas the frame
conditions specify what is not changed by the action. Axioms for the predicate
$enabled$ specify the preconditions of actions.

The formal semantics of a Hoare triple for conditional actions is derived from
the semantics of a GOAL agent and is defined relative to the set of traces
$S_A$ associated with the GOAL agent $A$. A Hoare triple for conditional
actions thus expresses a property of an agent and not just a property of an
action. The semantics of the basic capabilities are assumed to be fixed,
however, and are not defined relative to an agent.

\begin{definition} 
\label{def:hoarsem} {\em (semantics of Hoare triples for basic actions)}\\
A {\em Hoare triple for basic capabilities} $\Hoare{\varphi}{\ac}{\psi}$
means that for all $\Sigma,\Gamma$
\begin{itemize}
\item $\mstate\models\varphi\wedge enabled(\ac)\Impl
	\update(\ac,\mstate)\models\psi$, and
\item $\mstate\models\varphi\wedge\neg enabled(\ac)\Impl
	\mstate\models\psi$.
\end{itemize}
\end{definition}

To explain this definition, note that we made a case distinction between
states in which the basic action is enabled and in which it is not enabled.
In case the action is enabled, the postcondition $\psi$ of the Hoare triple
$\Hoare{\varphi}{\ac}{\psi}$ should be evaluated in the next state resulting
from executing action $\ac$. In case the action is not enabled, however,
the postcondition should be evaluated in the same state because a failed
attempt to execute action $\ac$ is interpreted as an idle step in which
nothing changes.

Hoare triples for conditional actions are interpreted {\em relative to the set
of traces} associated with the GOAL agent of which the action is a part.
Below, we write $\varphi[s_i]$ to denote that a mental state formula
$\varphi$ holds in state $s_i$.

\begin{definition} {\em (semantics of Hoare triples for conditional actions)}\\
\label{def:semhoare}
Given an agent $A$,
a {\em Hoare triple for conditional actions} $\Hoare{\varphi}{b}{\psi}$
(for $A$) means that for all traces $s\in S_A$ and $i$, we have that
\[
(\varphi[s_i]\wedge b=b_i\in s)\Impl \psi[s_{i+1}]
\]
where $b_i\in s$ means that action $b_i$ is taken in state $i$ of trace $s$.
\end{definition}

Of course, there is a relation between the execution of basic actions and
that of conditional actions, and therefore there also is a relation between
the two types of Hoare triples. The following lemma makes this relation
precise.

\begin{lemma}\label{lem_Hoare}
Let $A$ be a GOAL agent and $S_A$ be the meaning of $A$.
Suppose that we have $\Hoare{\varphi\wedge\psi}{\ac}{\varphi'}$ and
$S_A\models(\varphi\wedge\neg\psi)\impl\varphi'$. Then we also have
$\Hoare{\varphi}{\psi\impl do(\ac)}{\varphi'}$.
\paragraph{Proof:}
We need to prove that $(\varphi[s_i]\wedge (\psi\impl do(\ac))=b_i\in s)\Impl
\varphi'[s_{i+1}]$. Therefore, assume
$\varphi[s_i]\wedge (\psi\impl do(\ac))=b_i\in s)$. Two cases need to be
distinguished: The case that the
condition $\psi$ holds in $s_i$ and the case that it does not hold in $s_i$.
In the former case, because we have $\Hoare{\varphi\wedge\psi}{\ac}{\varphi'}$
we then know that $s_{i+1}\models\varphi'$. In the latter case, the conditional
action is not executed and $s_{i+1}=s_i$. From
$((\varphi\wedge\neg\psi)\impl\varphi')[s_i]$, $\varphi[s_i]$ and
$\neg\psi[s_i]$ it then follows that
$\varphi'[s_{i+1}]$ since $\varphi'$ is a state formula.
\end{lemma}

The definition of Hoare triples presented here formalises a {\em total
correctness property}. A Hoare triple $\Hoare{\varphi}{b}{\psi}$ ensures
that if initially $\varphi$ holds, then an attempt to execute $b$ results
in a successor state and in that state $\psi$ holds. This is different
from {\em partial correctness} where no claims about the termination of
actions and the existence of successor states are made.

\subsection{Basic Action Theories}
A {\em basic action theory} specifies the effects of the basic capabilities
of an agent. It specifies when an action is enabled, it specifies the effects
of an action and what does not change when an action is executed. Therefore,
a basic action theory consists of axioms for the predicate $enabled$ for each
basic capability, Hoare triples that specify the effects of basic capabilities
and Hoare triples that specify frame axioms associated with these capabilities.
Since the belief update capabilities of an agent are not fixed by the language
GOAL but are user-defined, the user should specify the axioms and Hoare triples
for belief update capabilities. The special actions for goal updating $\adopt$
and $\drop$ are part of GOAL and a set of axioms and Hoare triples for these
actions is specified below.

\subsubsection{Actions on beliefs: capabilities of the shopping assistant}
Because in this paper, our concern is not with the specification of basic
action theories in particular, but with providing a programming framework for
agents in which such specifications can be plugged in, we only provide some
example specifications of the capabilities defined in the personal assistant
example that we need in the proof of correctness below.

First, we specify a set of axioms for each of our basic actions that state
when that action is enabled. Below, we abbreviate the book titles of the
example, and write $T$ for {\em The Intentional Stance} and $I$ for {\em
Intentions, Plans, and Practical Reason}. In the shopping agent example, we
then have:
\[
enabled(goto\_website(site))\bimpl\true,\\
enabled(search(book))\bimpl\bel(current\_website(Amazon.com)),\\
enabled(put\_in\_shopping\_cart(book))\bimpl
	\bel(current\_website(book)),\\
enabled(pay\_cart)\bimpl\\
\ \ \ ((\bel in\_cart(T)\vee\bel in\_cart(I))
 \wedge\bel current\_website(ContentCart)).
\]

Second, we list a number of effect axioms that specify the effects of a
capability in particular situations defined by the preconditions of the
Hoare triple.

\begin{itemize}
\item The action $goto\_website(site)$ results in moving to the relevant
	web page:\\
	$\Hoare{\true}{goto\_website(site)}{\bel current\_website(site)}$,
\item At Amazon.com, searching for a book results in finding
	a page with relevant information about the book:\\
	$\Hoare{\bel current\_website(Amazon.com)}
		{search(book)}
		{\bel current\_website(book)}$
\item On the page with information about a particular book, selecting the
	action\\
	$put\_in\_shopping\_cart(book)$ results in the book being put
	in the cart; also, a new web page appears on which the
	contents of the cart are listed:\\
	$
	\HoareV{\bel current\_website(book)}
		{put\_in\_shopping\_cart(book)}
		{\bel(in\_cart(book)\wedge current\_website(ContentCart))}
	$
\item In case $book$ is in the cart, and the current web page presents a list
	of all the books in the cart, the action $pay\_cart$ may be selected
	resulting in the buying of all listed books:\\
	$
	\HoareV{\bel(in\_cart(book)\wedge current\_website(ContentCart))}
		{pay\_cart}
		{\neg\bel in\_cart(book)\wedge
		\bel(bought(book)\wedge current\_website(Amazon.com))}
	$
\end{itemize}

Finally, we need a number of frame axioms that specify which properties are
not changed by each of the capabilities of the agent. For example, both the
capabilities $goto\_website(site)$ and $search(book)$ do not change any
beliefs about $in\_cart$. Thus we have, e.g.:
\[
\Hoare{\bel in\_cart(book)}{goto\_website(site)}{\bel in\_cart(book)}\\
\Hoare{\bel in\_cart(book)}{search(book)}{\bel in\_cart(book)}
\]
It will be clear that we need more frame axioms than these two, and some of
these will be specified below in the proof of the correctness of the
shopping agent.

It is important to realise that the only Hoare triples that need to be
specified for agent capabilities are Hoare triples that concern the
effects upon the {\em beliefs} of the agent. Changes and persistence of
(some) goals due to executing actions can be derived with the proof rules
and axioms below that are specifically designed to reason about the effects
of actions on goals.

\subsubsection{Actions on goals}
A theory of the belief update capabilities and their effects on the beliefs
of an agent must be complemented with a theory about the effects of actions
upon the goals of an agent. Such a theory should capture both the effects of
the default commitment strategy as well as give a formal specification of the
the $\drop$ and $\adopt$ actions. Only in Section~\ref{subsec:complete}
we aim at providing a complete system, in the discussion in the
current
section, there are dependencies between the axioms and rules discussed.

\paragraph*{Default commitment strategy}
The default commitment strategy imposes a constraint on the persistence of
goals. A goal persists if it is not the case that after doing $\ac$ the
goal is believed to be achieved. Only action $\drop(\phi)$ is allowed to
overrule this constraint. Therefore, in case $\ac\neq\drop(\phi)$, we have
that $\Hoare{\goal\phi}{\ac}{\bel\phi\vee\goal\phi}$ (using 
the rule for conditional actions from Table~\ref{table:conditional},
one can derive that this triple also holds for general conditional
actions $b$, rather than just actions $\ac$). The
Hoare triple precisely captures the default commitment strategy and states
that after executing an action the agent either believes it has achieved
$\phi$ or it still has the goal $\phi$ if $\phi$ was a goal initially.

\begin{table}[!h]
\begin{center}
\framebox[7.5cm]{
\transro{\ac\neq\drop(\phi)}{\Hoare{\goal\phi}{\ac}{\bel\phi\vee\goal\phi}}
}
\caption{Persistence of goals}
\label{table:persistenceofgoals}
\end{center}
\end{table}

A similar Hoare triple can be given for the persistence of the absence of
a goal. Formally, we have
\begin{equation}
\label{eq:hoare1}
\Hoare{\neg\goal\phi}{b}{\neg\bel\phi\vee\neg\goal\phi}
\end{equation}

This Hoare triple
states that the absence of a goal $\phi$ persists, and in case it does not
persist the agent does not believe $\phi$ (anymore). The adoption of a goal
may be the result of executing an $\adopt$ action, of course. However, it
may also be the case that an agent believed it achieved $\phi$ but after
doing $b$ no longer believes this to be the case and adopts $\phi$ as a
goal again. For example, if the goal base is $\{p\wedge q\}$ and the
belief base $\bbase=\{p\}$, then the agent does not have a goal to achieve
$p$ because it already believes $p$ to be the case; however, in case an
action changes the belief base such that $p$ is no longer is believed, the
agent has a goal to achieve $p$ (again). This provides for a mechanism similar
to that of maintenance
goals. We do not need the Hoare triple~(\ref{eq:hoare1}) as an axiom, however, since it is a
direct consequence of the fact that $\bel\phi\impl\neg\goal\phi$ (this 
is exactly the postcondition of (\ref{eq:hoare1}). Note that
the stronger $\Hoare{\neg\goal\phi}{b}{\neg\goal\phi}$ does not hold, even if
$b\neq\varphi\impl do(\adopt(\phi))$. This occurs for example 
if we have $\goal(p \land q) \land \bel p$. Then the agent does
not have $p$ as a goal, since he believes it has already been achieved,
but, if he would give up $p$ as a belief, it becomes to be a goal.

In the semantics of Hoare triples (Definition~\ref{def:semhoare})
we stipulated that if $\ac$ is not enabled, we verify the postcondition
in the same state as the pre-condition:

\begin{table}[hb]
\begin{center}
\framebox[7.5cm]{
{$
	\transru
	{\varphi\impl\neg enabled(\ac)}
	{\Hoare{\varphi}{\ac}{\varphi}}
	$}
}
\caption{Infeasible actions}
\label{table:default}
\end{center}
\end{table}

\paragraph*{Frame properties on Beliefs}
The specification of the special actions $\drop$ and $\adopt$ involves a
number of frame axioms and a number of proof rules. The frame axioms capture
the fact that neither of these actions has any effect on the beliefs of an
agent. Note that, combining such properties with e.g. the
Consequence Rule (Table~\ref{table:structural}) one can derive the
triple $\transroax{}{\Hoare{\bel\psi}{\adopt(\phi)}{\neg\goal\psi}}$.

\begin{table}[h]
\begin{center}
\fbox{
\begin{minipage}{10.5cm}
\begin{tabbing}
$\Hoare{\bel\phi}{\adopt(\psi)}{\bel\phi}$ \ \ \=
	$\Hoare{\neg\bel\phi}{\adopt(\psi)}{\neg\bel\phi}$\\
$\Hoare{\bel\phi}{\drop(\psi)}{\bel\phi}$ \>
	$\Hoare{\neg\bel\phi}{\drop(\psi)}{\neg\bel\phi}$
\end{tabbing}
\end{minipage}}
\end{center}
\caption{Frame Properties on Beliefs for $\adopt$ and $\drop$}
\label{table:frame}
\end{table}

\paragraph*{(Non-)effects of $\adopt$}
The proof rules for the actions $\adopt$ and $\drop$ capture the effects
on the goals of an agent. For each action, we list proof rules for the
effect and the persistence (`non-effect') 
on the goal base for adoption (Table~\ref{table:adopt}) and
dropping (Table~\ref{table:drop})
of goals, respectively.
 
An agent adopts a new goal $\phi$ in case the agent does not believe
$\phi$ and $\phi$ is not a contradiction.
Concerning persistence, 
an adopt action does not remove any current goals of the agent. Any existing
goals thus persist when $\adopt$ is executed. The persistence of the absence
of goals is somewhat more complicated in the case of an $\adopt$ action. An
$\adopt(\phi)$ action does not add a new goal $\psi$ in case $\psi$ is not
entailed by $\phi$ or $\psi$ is believed to be the case:

\begin{table}[h]
\begin{center}
\fbox{
\begin{tabular}{ll}
  Effects of $\adopt$ &  \\
 \multicolumn{2}{c}{\transro{\not\cmodels\neg\phi}
	{\Hoare{\neg\bel\phi}{\adopt(\phi)}{\goal\phi}}}  \\
  Non-effect of $\adopt$ &  \\
$\transroax{}{\Hoare{\goal\psi}{\adopt(\phi)}{\goal\psi}}$
&
$\transro{\not\cmodels\psi\impl\phi}
	{\Hoare{\neg\goal\phi}{\adopt(\psi)}{\neg\goal\phi}}$

\end{tabular}
}
\end{center}
\caption{(Non-)effects of $\adopt$}
\label{table:adopt}
\end{table}

\weg{
\begin{table}[h]
\begin{center}
\framebox[7.5cm]{
\transro{\not\cmodels\neg\phi}
	{\Hoare{\neg\bel\phi}{\adopt(\phi)}{\goal\phi}}
}
\caption{The effect of $\adopt$}
\label{table:adopt}
\end{center}
\end{table}
}

\weg{
\begin{table}[h]
\begin{center}
\fbox{
\begin{tabular}{lll}
$\transroax{}{\Hoare{\goal\psi}{\adopt(\phi)}{\goal\psi}}$
&
$\transro{\not\cmodels\psi\impl\phi}
	{\Hoare{\neg\goal\phi}{\adopt(\psi)}{\neg\goal\phi}}$
&
$\transroax{}{\Hoare{\bel\psi}{\adopt(\phi)}{\neg\goal\psi}}$
\end{tabular}
}
\caption{Non-effects of $\adopt$}
\label{table:nonadopt}
\end{center}
\end{table}
}

A drop action $\drop(\phi)$ results in the removal of all goals that entail
$\phi$. This is captured by the first proof rule in Table~\ref{table:drop}
\weg{
\begin{table}[h]
\begin{center}
\framebox[7.5cm]{
\transro{\cmodels\psi\impl\phi}
	{\Hoare{\goal\psi}{\drop(\phi)}{\neg\goal\psi}}
}
\caption{Effect of $\drop$}
\label{table:drop}
\end{center}
\end{table}
}

\begin{table}[ht]
\begin{center}
\fbox{
\begin{tabular}{ll}
  Effects of $\drop$ &  \\
\multicolumn{2}{c}{$\transro{\cmodels\psi\impl\phi}
	{\Hoare{\goal\psi}{\drop(\phi)}{\neg\goal\psi}}$} \\
Non-Effects of $\drop$ & \\
$\transroax{}{\Hoare{\neg\goal\phi}{\drop(\psi)}{\neg\goal\phi}}$
&
$\transroax{}{\Hoare{\neg\goal(\phi\wedge\psi)\wedge\goal\phi}
	{\drop(\psi)}{\goal\phi}}$\\
\end{tabular}
}
\end{center}
\caption{(Non-)effects of $\drop$}
\label{table:drop}
\end{table}

Concerning persistence of goals under $\drop$:
a drop action $\drop(\phi)$ never results in the adoption of new goals.
The absence of a goal $\psi$ thus persists when a drop action is executed.
It is more difficult to formalise the persistence of a goal with respect to
a drop action. Since a drop action $\drop(\phi)$ removes goals which entail
$\phi$, to conclude that a goal $\psi$ persists after executing the action,
we must make sure that the goal does not depend on a goal (is a subgoal) that
is removed by the drop action. In case the conjunction $\phi\wedge\psi$ is
not a goal, we know this for certain.

\weg{
\begin{table}[h]
\begin{center}
\fbox{
$\transroax{}{\Hoare{\neg\goal\phi}{\drop(\psi)}{\neg\goal\phi}}
\ \ \ 
\transroax{}{\Hoare{\neg\goal(\phi\wedge\psi)\wedge\goal\phi}
	{\drop(\psi)}{\goal\phi}}$
}
\caption{Non-effects of $\drop$}
\label{table:nondrop}
\end{center}
\end{table}
}

The basic action theories for GOAL include a number of proof rules to derive
Hoare triples. The Rule for Infeasible Actions (Table~\ref{table:default})
allows to derive frame 
axioms for an action in case it is not enabled in a particular situation.
The Rule for Conditional Actions allows the derivation of Hoare triples for
conditional actions from Hoare triples for capabilities. This rule is
justified by lemma \ref{lem_Hoare}. Finally, there are three rules for
combining Hoare triples and for strengthening the precondition and weakening
the postcondition: they are displayed in Table~\ref{table:structural}.

Finally, we list how one goes from simple actions $\ac$ to conditional
actions $b$, and how to use complex formulas in pre- and post-conditions.

\begin{table}[h]
\begin{center}
\framebox[7.5cm]{
{$
	\transru
	{\Hoare{\varphi\wedge\psi}{{\sf a}}{\varphi'},
	(\varphi\wedge\neg\psi)\impl\varphi'}
	{\Hoare{\varphi}{\psi\arr do({\sf a})}{\varphi'}}
	$}
}
\caption{Rule for conditional actions}
\label{table:conditional}
\end{center}
\end{table}

We did not aim in this section at giving a weakest set of rules: in fact, the
rules that manipulate the pre- and post-conditions 
in Table~\ref{table:structural} for conditional actions $b$
could already be derived if we had only given them from simple actions $\ac$.
\begin{table}[h]
\begin{center}
\fbox{
\begin{tabular}{ll}
Consequence Rule:
& Conjunction Rule:\\
	{$
	\transru
	{\varphi'\impl\varphi, \Hoare{\varphi}{\ac}{\psi}, \psi\impl\psi' }
	{\Hoare{\varphi'}{\ac}{\psi'}}
	$}
& 
	{$
	\transru
	{\Hoare{\varphi_1}{b}{\psi_1}, \Hoare{\varphi_2}{b}{\psi_2}}
	{\Hoare{\varphi_1\wedge\varphi_2}{b}{\psi_1\wedge\psi_2}}
	$}\\
&\\
\multicolumn{2}{c}{Disjunction Rule:}\\
\multicolumn{2}{c}{$
	\transru
	{\Hoare{\varphi_1}{b}{\psi}, \Hoare{\varphi_2}{b}{\psi}}
	{\Hoare{\varphi_1\vee\varphi_2}{b}{\psi}}
	$}
\end{tabular}}
\caption{Structural rules}
\label{table:structural}
\end{center}
\end{table}

\subsection{A Complete Hoare System}
\label{subsec:complete}
We now address the issue of finding a complete Hoare system for
GOAL. 
Let $\hvdash \Hoare{\precond}{S}{\postcond}$ denote that the Hoare triple
with pre-condition $\precond$ and postcondition $\postcond$ is derivable 
in the calculus $H$ that we are about to introduce, and let $\hmodels$
denote the truth of such assertions, that is, $\hmodels$ determines
the truth of mental state formulas (Definition~\ref{def:mentalsem}),
that of formulas with $\enabled(-)$ (Definition~\ref{def:enabledsem})
and that of Hoare triples (Definition~\ref{def:hoarsem}), 
on mental states $\mstate$. From now on, the statement $S$
ranges over basic actions ${\sf a}$ and conditional actions $b$.
Then, this subsection wants to settle whether our calculus $H$ is
sound and complete, i.e.\ whether it can be proven that, for any 
pre- and postcondition $\precond$ and $\postcond \in \La_M$,

\[
\hvdash \Hoare{\precond}{S}{\postcond} \ \Longleftrightarrow \ 
\hmodels \Hoare{\precond}{S}{\postcond}
\]

Finding such a complete system is, even for 
`ordinary deterministic programs', impossible, since such
programs are interpreted over domains that include the integers,
and by G\"odel's Incompleteness Theorem, we know that axiomatize
a domain that includes those integers (cf.\ \cite{aptolderog}).
Here, our domain is not that of the integers, but, instead, we
will assume a completeness result for the basic capabilities {\em Bcap}
that modify the belief base, so that we can concentrate on the
actions that modify the goals. 

\begin{definition} {\em (General Substitutions)}\\
We define the following general substitution scheme. 
Let $\postcond, \alpha, \beta$ be mental state formulas, and
$X$ a variable ranging over formulas from $\La$. Then $\beta(X)$ 
denotes that $X$ may occur in $\beta$. Let $C(X)$ denote
a condition on $X$.
Then $\postcond[\alpha / \beta(X) |C(X)]$ denotes the 
formula that is obtained from $\postcond$ by substituting all
occurrences in $\postcond$ of $\beta(X)$ for which $C(X)$ holds,
by $\alpha$.
\end{definition}

For instance, the result of 
$(\goal{p} \land \neg \goal{q} \land \goal{s})[\bel{r}/ \goal{\phi} | 
(p \land q) \rightarrow \phi]$ is $\bel{r} \land \neg \bel{r} \land \goal{s}$.

\begin{definition}\label{def:hsystem}(The system H)\\
The valid Hoare triples for GOAL are as follows:\\
\begin{tabular}{ll}
{\sc Belief Capabilities} & $\Hoare{\precond(\ac)}{\ac}{\postcond(\ac)}$\\
{\sc Adopt}               & $\{(\enabled(\adopt(\phi)) 
\land \postcond[\neg\bel\phi'/ \goal\phi' |\cvdash \phi \rightarrow \phi'])$\\
& $\lor (\neg\enabled(\adopt(\phi)) \land \postcond)\}
\{\adopt(\phi)\}\{\postcond\}$\\
{\sc Drop} & $\Hoare{\postcond[\true/\neg\goal{\phi'}|\ \cvdash \phi' \rightarrow \phi]}
{\drop{(\phi)}}{\postcond}$\\
 & \\
{\sc Conditional Actions} & $
	\transru
	{\Hoare{\precond\wedge\psi}{{\sf a}}{\postcond},
	\models_{ME} (\postcond\wedge\neg\psi)\impl\precond}
	{\Hoare{\precond}{\psi\arr do({\sf a})}{\postcond}}
	$\\
  &  \\
{\sc Consequence Rule} & $\transru
	{\models_{ME} \postcond\impl\varphi, \Hoare{\varphi}{\ac}{\psi}, 
  \models_{ME}\psi\impl\postcond }
	{\Hoare{\precond}{\ac}{\postcond}}$\\
\end{tabular}
\end{definition}

\begin{lemma}\label{lem:subst}(Substitution Lemma)\\
\begin{enumerate}
\item[$(i)$]
Let $\mstate \mmodels \enabled(\adopt(\phi))$ and let
$\mstateprime = \update(\adopt(\phi),\mstate)$. Then: 
\[ \mstate \mmodels \postcond[\neg\bel\phi'/ \goal{\phi'} | \cvdash \phi \rightarrow \phi']  
\Longleftrightarrow \mstateprime \mmodels \postcond 
\]
\item[$(ii)$]
Let $\mstateprime = \update(\drop(\phi),\mstate)$. Then: 
\[
\mstate \mmodels \postcond[\true/\neg\goal{\phi'}|\ \cvdash \phi' \rightarrow \phi]
\Longleftrightarrow \mstateprime \mmodels \postcond 
\]
\end{enumerate}
\end{lemma}
{\bf Proof}. We only prove $(i)$, the proof of
$(ii)$ is similar. 
We prove $(i)$ by induction on 
the mental state formula $\postcond$.
\begin{enumerate}
\item
$\postcond$ is of the form $\goal\chi$. We distinguish two cases. 
\begin{enumerate}
\item
$\cvdash \phi \rightarrow \chi$. By definition of $\update(\adopt(\phi),\mstate)$,
we immediately see that $\mstate \mmodels \neg\bel\chi$ iff $\mstateprime \mmodels \goal\chi$.
\item
$\not\cvdash \phi \rightarrow \chi$. The definition of $\update(\adopt(\phi),\mstate)$
guarantees that the adopt has no effect on the fact that
$\chi$ is a goal, thus we have $\mstate \hmodels \goal\chi$ iff
$\update(\adopt(\phi),\mstate) \hmodels \goal\chi$.
Also, the substitution has no change as an effect: 
$(\goal\chi)[\neg\bel\phi'/ \goal{\phi'} | \cvdash \chi \rightarrow \phi'] = \goal\chi$,
and we have the desired result.
\end{enumerate}
\item
$\postcond$ if of the form $\bel\chi$. In this case, the substitution had no effect,
and also, since the $\adopt$ has no effect on the belief base, we have that
$\mstate \mmodels \bel\chi$ iff $\mstateprime \mmodels \bel\chi$.
\item
The cases that $\postcond$ is a negation or a conjunction of mental states, 
follows immediately.
\end{enumerate}

\begin{lemma}(Soundness of $\hvdash$)\\
for any 
pre- and postcondition $\precond$ and $\postcond \in \La_M$,
$\hvdash \Hoare{\precond}{S}{\postcond} \ \Longrightarrow \ 
\hmodels \Hoare{\precond}{S}{\postcond}$
\end{lemma}
{\bf Proof}. 
Soundness of 
{\sc Belief Capabilities} is assumed. 
The cases for $\adopt$ and $\drop$ immediately follow
from Lemma~\ref{lem:subst}. The soundness of 
{\sc Consequence Rule} is easily proven using Theorem~\ref{thm:soundmvdash}.
Let us finally consider the rule {\sc Conditional Actions}.
Suppose that $\hmodels \Hoare{\postcond\wedge\psi}{{\sf a}}{\precond}$ (1),
and $\models_{ME}(\precond\wedge\neg\psi)\impl\postcond$ (2), and take an
arbitrary mental state $\mstate$. We have to demonstrate that
$\mstate \models_{ME} \Hoare{\precond}{\psi\arr do({\sf a})}{\postcond}$ (3).
Hence, we assume that $\mstate \models_{ME} \precond$. We then distinguish
two cases. First, assume $\mstate \models_{ER} \psi$.
Then, by our assumption (1), we have that $\mstateprime \mmodels \postcond$,
for $\mstateprime = \update({\sf a},\mstate)$. The second case is the one
in which $\mstate \not\mmodels \psi$, i.e., $\mstate\mmodels\neg\psi$. 
By (2), we know that $\mstateprime \models_{ER} \postcond$, for 
any $\mstateprime$  for which $\langle\bbase,\gbase\rangle\stackrel{b}{\lra}
	\langle\bbase',\gbase'\rangle$. All in all, 
we have proven (3).

We first introduce the notion
of weakest liberal precondition, originally due to Dijkstra (\cite{dijkstra}).
However, we introduce it immediately in the syntax.

\begin{definition}(Weakest Liberal Precondition)\\
For $S = \ac', \adopt(\phi) \drop(\phi)$ and $\psi \rightarrow \ac$, 
($\ac'$ a belief capability, $\ac$ any basic capability), we
define the weakest liberal precondition for $S$ to achieve $\postcond$,
$\wlp(S,\postcond) \in \La_m$, as follows:
\begin{enumerate}
\item
If $\ac'$ is a belief capability, and 
$\hvdash \Hoare{\postcond^{-1}(\ac')}{\ac'}{\postcond(\ac')}$
is the rule for $\ac'$, then $\wlp(\ac',\postcond(\ac')) = \postcond^{-1}$
\item
$\wlp(\adopt(\phi),\postcond) = 
(\enabled(\adopt(\phi)) 
\land \postcond[\neg\bel\phi'/ \goal\phi' |\cvdash \phi \rightarrow \phi')$\\
$\lor (\neg\enabled(\adopt(\phi)) \land \postcond)$
\item
$\wlp(\drop(\phi),\postcond) = 
\postcond[\true/\neg\goal{\phi'}|\ \cvdash \phi' \rightarrow \phi]$
\item
$\wlp(\psi \rightarrow \ac,\postcond) = (\psi \land \wlp(\ac,\postcond))
\lor (\neg\psi \land \postcond)$
\end{enumerate}
\end{definition}

Note that the weakest precondition $\wlp(S,\postcond)$ is 
indeed an mental state formula. 

\begin{lemma}\label{lem:wlp}(Weakest Precondition Lemma)\\
We have: $\hvdash \Hoare{\wlp(S,\postcond}{S}{\postcond}$, for 
every $S$ and every postcondition $\postcond$.
\end{lemma}
{\bf Proof}. For $S = \ac', \adopt(\phi), \drop(\phi)$, this 
follows immediately from Definition~\ref{def:hsystem}. For 
conditional actions $b = \psi \rightarrow \ac$, 
we have to prove 
\[\hvdash \Hoare{(\psi \land \wlp(\ac,\postcond))
\lor (\neg\psi \land \postcond)}{\psi \rightarrow \ac}{\postcond}\]
Let us abbreviate $(\psi \land \wlp(\ac,\postcond))
\lor (\neg\psi \land \postcond)$ to $\postcond^{-1}_b$.
The induction hypothesis tells us
that $\hvdash \Hoare{\wlp(\ac,\postcond)}{\ac}{\postcond}$ (1).
Note that $\mmodels \postcond^{-1}_b \rightarrow \wlp(\ac,\postcond)$.
Hence, by the {\sc Consequence Rule} and (1), we have
$\hvdash \Hoare{\postcond^{-1}_b}{\ac}{\postcond}$ (2).
Obviously, we also have $\mmodels (\postcond^{-1}_b \land \neg \psi) \rightarrow
\postcond$ (3). Applying the {\sc Conditional Actions} rule to (2) 
and (3), we conclude 
$\hvdash \Hoare{\postcond^{-1}_b}{\psi \rightarrow \ac}{\postcond}$,
which was to be proven.

\begin{lemma}\label{lem:posttowlp}\ \\
$\hmodels \Hoare{\precond}{S}{\postcond} \Rightarrow 
\mmodels \precond \rightarrow \wlp(S,\postcond)$
\end{lemma}
{\bf Proof}.
We even prove a stronger statement, i.e., that for all
mental states $\mstate$, if $\mstate \hmodels \Hoare{\precond}{S}{\postcond}$,
then $\mstate \mmodels \precond \rightarrow \wlp(S,\postcond)$.
To prove this, we take an arbitrary $\mstate$ for which both
$\mstate \mmodels \precond$ and 
$\mstate \hmodels \Hoare{\precond}{S}{\postcond}$
and we then have to show that $\mstate \mmodels \wlp(S,\postcond)$.
\begin{enumerate}
\item
$S = \adopt(\phi)$. We know that $\mstate \hmodels 
\Hoare{\precond}{\adopt(\phi)}{\postcond}$. We distinguish two
cases, the first of which says that 
$\mstate \mmodels \neg\enabled(\adopt(\phi))$.
Then we have a transition from $\mstate$ to itself, and hence
$\mstate \mmodels \postcond$, and hence 
$\mstate \mmodels \neg\enabled(\adopt(\phi)) \land \postcond$, so that
$\mstate \mmodels \wlp(\adopt(\phi),\postcond)$. In the second case,
$\mstate \mmodels \enabled(\adopt(\phi))$. 
By the Substitution Lemma~\ref{lem:subst}, case $(i)$, we
then immediately see $\mstate 
\mmodels \postcond[\neg\bel\phi'/ \goal{\phi'} | 
\cvdash \phi \rightarrow \phi']$, and hence 
$\mstate \mmodels \wlp(\adopt(\phi,\postcond)$.
\item
$S = \drop(\phi)$. This case follows immediately from the Substitution Lemma
and the definition
of $\wlp(\drop(\phi,\postcond)$.
\item
$S = \psi \rightarrow \ac$. 
We know that $\mstate \hmodels 
\Hoare{\precond}{\psi \rightarrow \ac}{\postcond}$ and
that $\mstate \mmodels \precond$. 
If $\mstate \mmodels \neg\psi$, then the transition belonging
to $S$ ends up in $\mstate$, and hence we then have 
$\mstate \mmodels \neg\psi \land \postcond$, and in 
particular $\mstate \mmodels \wlp(S,\postcond)$.
In the other case we have $\mstate \mmodels \psi$, and
we then know that $\mstate \hmodels \Hoare{\precond}{\ac}{\postcond}$.
By induction we conclude that $\mstate \mmodels \wlp(\ac,\postcond$,
and hence $\mstate \mmodels \wlp(\psi \rightarrow \ac),\postcond)$.
\end{enumerate}

\begin{theorem}(Completeness of $\hvdash$))\\
For any 
pre- and postcondition $\precond$ and $\postcond \in \La_M$,

\[
\hvdash \Hoare{\precond}{S}{\postcond} \ \Longleftrightarrow \ 
\hmodels \Hoare{\precond}{S}{\postcond}
\]
\end{theorem}

{\bf Proof}.
Suppose $\mmodels \Hoare{\precond}{S}{\postcond}$. 
The Weakest Precondition Lemma (\ref{lem:wlp})
tells us that 
$\hvdash \Hoare{\wlp(S,\postcond)}{S}{\postcond}$ (1). 
By Lemma~\ref{lem:posttowlp}, we have 
$\mmodels \postcond \rightarrow \wlp(S,\postcond)$ (2).
We finally apply the {\sc Consequence Rule} to (1) and (2) 
to conclude to conclude that 
$\hvdash \Hoare{\postcond}{S}{\precond}$.

\subsection{Temporal logic}
\label{subsec:temporal}
On top of the Hoare triples for specifying actions, a temporal logic
is used to specify and verify properties of GOAL agents. Two new operators
are introduced. The proposition $\initc$ states that the agent is at the
beginning of execution and nothing has happened yet. The second operator
$\until$ is a weak until operator. $\varphi\until\psi$ means that
$\psi$ eventually becomes true and $\varphi$ is true until $\psi$ becomes
true, or $\psi$ never becomes true and $\varphi$ remains true forever.

\begin{definition} {\em (language of temporal logic $\La_T$ based on $\La$)}\\
The {\em temporal logic language $\La_T$} is inductively defined by:
\begin{itemize}
\item $\initc\in\La_T$,
\item $enabled(\ac), enabled(\varphi\impl do(\ac))\in\La_T$ for $\ac\in Cap$,
\item if $\phi\in\La$, then $\bel\phi,\goal\phi\in\La_T$,
\item if $\varphi,\psi\in\La_T$, then $\neg\varphi,\varphi\wedge\psi\in\La_T$,
\item if $\varphi,\psi\in\La_T$, then $\varphi\until\psi\in\La_T$.
\end{itemize}
\end{definition}

A number of other well known temporal operators can be defined in terms
of the operator $\until$. The {\em always} operator $\Box\varphi$ is an
abbreviation for $\varphi\until\false$, and the {\em eventuality} operator
$\diamond\varphi$ is defined as $\neg\Box\neg\varphi$ as usual.

Temporal formulas are evaluated with respect to a trace $s$ and a time point
$i$. State formulas like $\bel\phi$, $\goal\psi$, $enabled(\ac)$ etc. are
evaluated with respect to mental states.

\begin{definition} {\em (semantics of temporal formulas)}\\
Let $s$ be a trace and $i$ be a natural number.
\begin{itemize}
\item $s,i\models\initc$ iff $i=0$,
\item $s,i\models enabled(\ac)$ iff $enabled(\ac)[s_i]$,
\item $s,i\models enabled(\varphi\impl do(\ac))$ iff
	$enabled(\varphi\impl do(\ac))[s_i]$,
\item $s,i\models\bel\phi$ iff
	$\bel\phi[s_i]$,
\item $s,i\models\goal\phi$ iff
	$\goal\phi[s_i]$,
\item $s,i\models\neg\varphi$ iff
	$s,i\not\models\varphi$,
\item $s,i\models\varphi\wedge\psi$ iff
	$s,i\models\varphi$ and
	$s,i\models\psi$,
\item $s,i\models\varphi\until\psi$ iff
	$\exists j\geq i(s,j\models\psi\wedge
		\forall k(i\leq k< j(s,k\models\varphi)))$ or
		$\forall k\geq i(s,k\models\varphi)$.
\end{itemize}
\end{definition}

We are particularly interested in temporal formulas that are valid with
respect to the set of traces $S_A$ associated with a GOAL agent $A$.
Temporal formulas valid with respect to $S_A$ express properties of the
agent $A$.

\begin{definition}
Let $S$ be a set of traces.
\begin{itemize}
\item $S\models\varphi$ iff $\forall s\in S, i(s,i\models\varphi)$,
\item $\models\varphi$ iff $S\models\varphi$ where $S$ is the set of
	all traces.
\end{itemize}
\end{definition}

In general, two important types of temporal properties can be distinguished.
Temporal properties are divided into {\em liveness} and {\em safety}
properties. Liveness properties concern the progress that a program makes
and express that a (good) state eventually will be reached. Safety properties,
on the other hand, express that some (bad) state will never be entered.
In the rest of this section, we discuss a number of specific liveness and
safety properties of an agent $A=\langle\Pi_A,\bbase_0,\gbase_0\rangle$.

We show that each of the properties that we discuss are equivalent to a set
of Hoare triples. The importance of this result is that it shows that
temporal properties of agents can be proven by inspection of the program
text only. The fact that proofs of agent properties can be constructed by
inspection of the program text means that there is no need to reason about
individual traces of an agent or its operational behaviour. In general,
reasoning about the program text is more economical since the number of
traces associated with a program is exponential in the size of the program.

The first property we discuss concerns a safety property, and is expressed
by the temporal formula $\varphi\impl(\varphi\until\psi)$. Properties in this
context always refer to agent properties and are evaluated with respect to the
set of traces associated with that agent. Therefore, we can explain the
informal meaning of the property as stating that if $\varphi$ ever becomes
true, then it remains true until $\psi$ becomes true. By definition, we
write this property as $\varphi\unless\psi$:
\[
\varphi\unless\psi \stackrel{df}{=}
\varphi\impl(\varphi\until\psi)
\]

An important special case of an $\unless$ property is $\varphi\unless\false$,
which expresses that if $\varphi$ ever becomes true, it will remain true.
$\varphi\unless\false$ means that $\varphi$ is a {\em stable} property of
the agent.
In case we also have $\initc\impl\varphi$, where $\initc$ denotes the initial
starting point of execution, $\varphi$ is always true and is an {\em invariant}
of the program.

Now we show that $\unless$ properties of an agent
$A=\langle\Pi,\sigma_0,\gamma_0\rangle$ are equivalent to a
set of Hoare triples for basic actions in $\Pi$. This shows that we can prove
$\unless$ properties by proving a finite 
set of Hoare triples. The proof relies on
the fact that if we can prove that after executing any action from $\Pi$
either $\varphi$ persists or $\psi$ becomes true, we can conclude that
$\varphi\unless\psi$.

\begin{theorem}\label{th_unless}
Let $A=\langle\Pi_A,\bbase_0,\gbase_0\rangle$. Then:
\[
\forall b\in\Pi_A(\Hoare{\varphi\wedge\neg\psi}{b}{\varphi\vee\psi})
\mbox{ iff }
S_A\models\varphi\unless\psi
\]
\paragraph{Proof:}
The proof from right to left is the easiest direction in the proof.
Suppose $S_A\models\varphi\unless\psi$ and $s,i\models\varphi$. This
implies that $s,i\models\varphi\until\psi$. In case we also have
$s,i\models\psi$, we are done. So, assume $s,i\models\neg\psi$ and action
$b$ is selected in the trace at state $s_i$. From the semantics of $\until$
we then know that $\varphi\vee\psi$ holds at state $s_{i+1}$, and we
immediately obtain $\Hoare{\varphi\wedge\neg\psi}{b}{\varphi\vee\psi}$
since $s$ and $i$ were arbitrarily chosen trace and time point. To prove
the Hoare triple for the other actions in the agent program $A$, note that
when we replace action $b$ with another action $c$ from $\Pi_A$ in trace $s$,
the new trace $s'$ is still a valid trace that is in the set $S_A$. Because
we have $S_A\models\varphi\unless\psi$, we also have
$s',i\models\varphi\unless\psi$ and from reasoning by analogy we obtain the
Hoare triple for action $c$ (and similarly for all other actions).

We prove the left to right case by contraposition. Suppose that
\[
(*)\; \forall b\in\Pi_A(\Hoare{\varphi\wedge\neg\psi}{b}{\varphi\vee\psi})
\]
and for some $s\in S_A$ we have $s,i\not\models\varphi\unless\psi$. The
latter fact means that we have $s,i\models\varphi$ and
$s,i\not\models\varphi\until\psi$. $s,i\not\models\varphi\until\psi$ implies
that either (i) $\psi$ is never established at some $j\geq i$ but we do have
$\neg\varphi$ at some time point $k>i$ or (ii) $\psi$ is established at some
time $j>i$, but in between $i$ and any such $j$ it is not always the case that
$\varphi$ holds.

In the first case (i), let $k>i$ be the smallest $k$ such that
$s,k\not\models\varphi$. Then, we have $s,{k-1}\models\varphi\wedge\neg\psi$
and $s,k\models\neg\varphi\wedge\neg\psi$. In state $s_{k-1}$, however, either
a conditional action is performed or no action is performed. From (*) we then
derive a contradiction.

In the second case (ii), let $k>i$ be the smallest $k$ such that
$s,k\models\psi$. Then we know that there is a smallest $j$ such that $i<j<k$
and $s,j\not\models\varphi$ ($j\neq i$ since $s,i\models\varphi$). This means
that we have $s,{j-1}\models\varphi\wedge\neg\psi$. However, in state $s_j$
either a conditional action is performed or no action is performed. From (*)
we then again derive a contradiction.
\end{theorem}

Liveness properties involve eventualities which state that some state will be
reached starting from a particular situation. To express a special class of
such properties, we introduce the operator $\varphi\ensures\psi$.
$\varphi\ensures\psi$ informally means that $\varphi$ guarantees the
realisation of $\psi$, and is defined as:
\[
\varphi\ensures\psi \stackrel{df}{=} \varphi\unless\psi\wedge
(\varphi\impl\eventual\psi)
\]

$\varphi\ensures\psi$ thus ensures that $\psi$ is eventually realised starting
in a situation in which $\varphi$ holds, and requires that $\varphi$ holds
until $\psi$ is realised. For the class of $\ensures$ properties, we can show
that these properties can be proven by proving a set of Hoare triples. The
proof of a $\ensures$ property thus can be reduced to the proof of a set of
Hoare triples.

\begin{theorem}\label{th_ensures}
Let $A=\langle\Pi_A,\sigma_0,\gamma_0\rangle$. Then:
\[
\forall b\in\Pi_A(\Hoare{\varphi\wedge\neg\psi}{b}{\varphi\vee\psi})\wedge
\exists b\in\Pi_A(\Hoare{\varphi\wedge\neg\psi}{b}{\psi})\\
\Impl S_A\models\varphi\ensures\psi 
\]
\paragraph{Proof:}
In the proof, we need the weak fairness assumption.
Since $\varphi\ensures\psi$ is defined as
$\varphi\unless\psi\wedge(\varphi\impl\eventual\psi)$, by theorem
\ref{th_unless} we only need to prove that
$S_A\models\varphi\impl\eventual\psi$ given that
$\forall b\in\Pi_A(\Hoare{\varphi\wedge\neg\psi}{b}{\varphi\vee\psi})\wedge
\exists b\in\Pi_A(\Hoare{\varphi\wedge\neg\psi}{b}{\psi})$.
Now suppose, to arrive at a contradiction, that for some time point $i$ and
trace $s\in S_A$ we have: $s,i\models\varphi\wedge\neg\psi$ and assume that
for all later points $j>i$ we have $s,j\models\neg\psi$. In that case, we
know that for all  $j>i$ we have $s,j\models\varphi\wedge\neg\psi$
(because we may assume $\varphi\unless\psi$). However, we also know that
there is an action $b$ that is enabled in a state in which
$\varphi\wedge\neg\psi$ holds and transforms this state to a state in
which $\psi$ holds. The action $b$ thus is always enabled, but apparently
never taken. This is forbidden by weak fairness, and we arrive at a
contradiction.
\end{theorem}

\weg{
The implication in the other direction in theorem \ref{th_ensures} does not
hold. A counterexample is provided by the program:
\[
\Pi=\{\\
\t1 \bel(\neg p\wedge q)\impl do(\myins(p)),\\
\t1 \bel(\neg p\wedge r)\impl do(\myins(p)),\\
\t1 \bel p\impl do(\myins(\neg p\wedge q)),\\
\t1 \bel p\impl do(\myins(\neg p\wedge r))\},\\
\bbase_0=\{ p\},\\
\gbase_0=\emptyset.
\]
where
\[ \Trans(\myins(p),\{\neg p\wedge q\})=\{p\wedge q\},
\Trans(\myins(p),\{\neg p\wedge r\})=\{p\wedge r\},\\
\Trans(\myins(\neg p\wedge q),\{p\})=
\Trans(\myins(\neg p\wedge q),\{\{p\wedge q\})=\{\neg p\wedge q\},\\
\Trans(\myins(\neg p\wedge r),\{p\})=
\Trans(\myins(\neg p\wedge r),\{p\wedge r\})=\{\neg p\wedge r\}.
\]
For this program, we have that $\bel\neg p\ensures\bel p$ holds, but we do not
have $\Hoare{\bel\neg p\wedge\neg\bel p}{b}{\bel p}$ for any $b\in \Pi$.
}

Finally, we introduce a third temporal operator `leads to' $\mapsto$. The
operator $\varphi\mapsto\psi$ differs from $\ensures$ in that it does not
require $\varphi$ to remain true until $\psi$ is established, and is
derived from the $\ensures$ operator. $\mapsto$
is defined as the transitive, disjunctive closure of $\ensures$.

\begin{definition} {\em (leads to operator)}\\
The leads to operator $\mapsto$ is defined by:
\begin{center}
\fbox{
\begin{tabular}{llll}
$
\transru{\varphi\ensures\psi}{\varphi\mapsto\psi}$
&
$\transru{\varphi\mapsto\chi, \chi\mapsto\psi}{\varphi\mapsto\psi}$&
$\transru{\varphi_1\mapsto\psi,\ldots,\varphi_n\mapsto\psi}
	{(\varphi_1\vee\ldots\vee\varphi_n)\impl\psi}$
\weg{&
$\transru{\varphi\impl\eventual\psi}{\varphi\mapsto\psi}$}
\end{tabular}
}\end{center}
\end{definition}

The meaning of the `leads to' operator is captured by the following lemma.
$\varphi\mapsto\psi$ means that given $\varphi$ condition $\psi$ will
eventually be realised. The proof of the lemma is an easy induction on
the definition of $\mapsto$.

\begin{lemma}
$\varphi\mapsto\psi\models\varphi\impl\eventual\psi$.
\end{lemma}

\section{Proving Agents Correct}
In this section, we use the programming logic to prove the correctness of
our example shopping agent. We do not present all the details, but provide
enough details to illustrate the use of the programming logic. Before we
discuss what it means that an agent
program is correct and provide a proof which shows that our example agent is
correct, we introduce some notation. The notation involves a number
of abbreviations concerning names and propositions in the language of our
example agent:
\begin{itemize}
\item Instead of $current\_website(sitename)$ we just write $sitename$;
	e.g., we write $Am.$$com$ and $ContentCart$ instead of
	$current\_website$$(Am.com)$ and $current\_website$$(ContentCart)$,
respectively.
\item As before, the book titles {\em The Intentional Stance} and
	{\em Intentions,
	Plans and Practical Reason} that the agent intends to buy are
	abbreviated to $T$ and $I$ respectively. These conventions
	can result in formulas like $\bel(T)$, which means that the agent
	is at the web page concerning the book {\em The Intentional Stance}.
\end{itemize}

A simple and intuitive correctness property, which is natural in this
context and is applicable to our example agent, states that a GOAL agent
is {\em correct} when the agent program realises the initial goals of the
agent. For this subclass of correctness properties, we may consider the agent
to be finished upon establishing the initial goals and in that case the agent
could be terminated. Of course, it is also possible to continue the execution
of such agents. This class of correctness properties can be expressed by means
of temporal formulas like $\goal\phi\impl\diamond\neg\goal\phi$. Other
correctness properties are conceivable, of course, but not all of them can
be expressed easily in the temporal proof logic for GOAL.

\subsection{Correctness Property of the Shopping Agent}

From the discussion above, we conclude that the interesting property to
prove for our example program is the following property:
\[
Bcond\wedge\goal(bought(T)\wedge bought(I))
\mapsto \bel(bought(T)\wedge bought(I))
\]
where $Bcond$ is some condition of the initial beliefs of the agent.
More specifically, $Bcond$ is defined by:
\[
\bel current\_webpage(hpage(user))\wedge
\neg\bel in\_cart(T)\wedge\neg\bel in\_cart(I)\wedge\\
\bel(\forall s,s'((s\neq s'\wedge current\_webpage(s))\impl
	\neg current\_webpage(s')))
\]
The correctness property states that the goal to buy the books
{\em The Intentional Stance} and {\em Intentions, Plans and Practical Reason},
given some initial
conditions on the beliefs of the agent, leads to buying (or believing to have
bought) these books.
Note that this property expresses a {\em total correctness} property. It
states both that the program behaves as desired and that it will eventually
reach the desired goal state. An extra reason for considering this property
to express correctness of our example agent is that the goals involved once
they are achieved remain true forever (they are `stable' properties).

\subsection{Invariants and Frame Axioms}

To be able to prove correctness, we need a number of frame axioms.
There is a close relation between frame axioms and invariants of a program.
This is because frame axioms express properties that are not changed by
actions, and a property that, once true, remains true whatever action is
performed is a stable property. In case such a property also holds initially,
the property is an {\em invariant} of the program. In our example program,
there is one invariant that states that it is impossible to be at two web
pages at the same time:
$inv=\bel\forall s,s'((s\neq s'\wedge current\_webpage(s))\impl
	\neg current\_webpage(s'))$.

To prove that $inv$ is an invariant of the agent, we need frame axioms
stating that when $inv$ holds before the execution of an action it still
holds after executing that action. Formally, for each $\ac\in Cap$,
we need: $\Hoare{inv}{\ac}{inv}$. These frame axioms need to be specified
by the user, and for our example agent we assume that they are indeed true.
By means of the Consequence Rule (strengthen the precondition of the Hoare
triples for capabilities $\ac$) and the Rule for Conditional Actions
(instantiate $\varphi$ and $\varphi'$ with $inv$), we then obtain that
$\Hoare{inv}{b}{inv}$ for all $b\in\Pi$. By theorem \ref{th_unless}, we
then know that $inv\unless\false$. Because we also have that initially
$inv$ holds since $\langle\sigma_0,\gamma_0\rangle\models inv$, we may
conclude that $\initc\impl\bel inv\wedge inv\unless\false$. $inv$ thus is
an invariant and holds at all times during the execution of the agent.
Because of this fact, we do not mention $inv$ explicitly anymore in the
proofs below, but will freely use the property when we need it.

A second property that is stable is the property $status(book)$:
\[
status(book)\stackrel{df}{=}
(\bel in\_cart(book)\wedge\goal bought(book))\vee\bel bought(book)
\]
The fact that $status(book)$ is stable means that once a book is in the
cart and it is a goal to buy the book, it remains in the cart and is only
removed from the cart when it is bought.

The proof obligations to prove that $status(book)$ is a {\em stable} property,
i.e. to prove that $status(book)\unless\false$, consist of supplying proofs for
\[
\Hoare{status(book)}{b}{status(book)}
\]
for each conditional action $b\in\Pi$
of the shopping agent (cf.\ theorem \ref{th_unless}). By the Rule for
Conditional Actions, therefore, it is sufficient to prove for each conditional
action $\psi\impl do(\ac)\in\Pi$ that
$\Hoare{status(book)\wedge\psi}{\ac}{status(book)}$ and
$(status(book)\wedge\neg\psi)\impl status(book)$. The latter implication is
trivial. Moreover, it is clear that to prove the Hoare triples it is sufficient
to prove $\Hoare{status(book)}{\ac}{status(book)}$ since we can strengthen
the precondition by means of the Consequence Rule. The proof obligations
thus reduce to proving $\Hoare{status(book)}{\ac}{status(book)}$ for each
capability of the shopping agent.

Again, we cannot prove these Hoare triples without a number of frame axioms.
Because no capability is allowed to reverse the fact that a book has been
bought, for each capability, we can specify a frame axiom for the predicate
$bought$:
\[
(1)\t1 \Hoare{\bel bought(book)}{\ac}{\bel bought(book)}
\]
In case the book is not yet bought, selecting action $pay\_cart$ may change
the contents of the cart and therefore we first treat the other three actions
$goto\_website$, $search$, and $put\_in\_shopping\_cart$ which are not supposed
to change the contents of the cart. For each of the latter three capabilities
we therefore add the frame axioms:
\[
\Hoare{\bel in\_cart(book)\wedge\neg\bel bought(book)}{\ac}
	{\bel in\_cart(book)\wedge\neg\bel bought(book)}
\]
where $\ac\neq pay\_cart$. Note that these frame axioms do not refer to goals
but only refer to the beliefs of the agent, in agreement with our claim that
only Hoare triples for belief updates need to be specified by the user. By
using the axiom $\goal bought(book)\impl\neg\bel bought(book)$ and the
Consequence Rule, however, we can conclude that:
\[
\Hoare{\bel in\_cart(book)\wedge\goal bought(book)}{\ac}
	{\bel in\_cart(book)\wedge\neg\bel bought(book)}
\]
By combining this with the axiom
\[
\Hoare{\goal bought(book)}{\ac}{\bel bought(book)\vee\goal bought(book)}
\]
by means of the Conjunction Rule and by rewriting the postcondition with
the Consequence Rule, we then obtain
\[
(2) \Hoare{\bel in\_cart(book)\wedge\goal bought(book)}{\ac}
	{\bel in\_cart(book)\wedge\goal bought(book)}
\]
where $\ac\neq pay\_cart$.
By weakening the postconditions of (1) and (2) by means of the Consequence
Rule and combining the result with the Disjunction Rule, it is then possible
to conclude that $\Hoare{status(book)}{\ac}{status(book)}$ for
$\ac\neq pay\_cart$.

As before, in the case of capability $pay\_cart$ we deal with each of the
disjuncts of $status(book)$ in turn. The second disjunct can be handled as
before, but the first disjunct is more involved this time because $pay\_cart$
can change both the content of the cart and the goal to buy a book if it is
enabled. Note that $pay\_cart$ only is enabled in case $\bel ContentCart$
holds. In case $\bel ContentCart$ holds and $pay\_cart$ is enabled, from the
effect axiom for $pay\_cart$ and the Consequence Rule we obtain

\begin{eqnarray*}
\HoareV{\bel in\_cart(book)\wedge\goal bought(book)\wedge\bel
	ContentCart}{\hspace{-2cm}(3)\hspace{2cm}pay\_cart}{\bel bought(book)}
\end{eqnarray*}
In case $\neg\bel ContentCart$ holds and $pay\_cart$ is not enabled, we
use the Rule for Infeasible Capabilities to conclude that
\begin{eqnarray*}
\HoareV{\bel in\_cart(book)\wedge\goal bought(book)\wedge
\neg\bel ContentCart}{\hspace{-2cm}(4)\hspace{2cm}pay\_cart}
{\bel in\_cart(book)\wedge\goal bought(book)\wedge \neg\bel ContentCart}
\end{eqnarray*}
By means of the Consequence Rule and the Disjunction Rule, we then can
conclude from (1), (3) and (4) that
$\Hoare{status(book)}{pay\_cart}{status(book)}$, and we are done.

\subsection{Proof Outline}

The main proof steps to prove our agent example correct are listed
next. The proof steps below consists of a number of $\ensures$ formulas
which together prove that the program reaches its goal in a finite number
of steps.\\

\noindent
\begin{tabular}{ll}
$(1)$&
$\bel hpage(user)\wedge\neg\bel in\_cart(T)\wedge\goal bought(T)\wedge$\\
& $\wedge\neg\bel in\_cart(I)\wedge \goal bought(I)$ \\
&$\ensures$\\
& $\bel Am.com\wedge\neg\bel in\_cart(T)\wedge\goal bought(T)\wedge
	\neg\bel in\_cart(I)\wedge\goal bought(I)$\\
\end{tabular}\mbox{ }\\ \vspace{3mm}
\begin{tabular}{ll}
$(2)$&
$\bel Am.com\wedge\neg\bel in\_cart(T)\wedge\goal bought(T)\wedge
	\neg\bel in\_cart(I)\wedge\goal bought(I)$\\
& $\ensures$\\
 & $[(\bel(T)\wedge\goal bought(T)\wedge\neg\bel in\_cart(I)\wedge
	\goal bought(I))\vee$\\
& 	$(\bel(I)\wedge\goal bought(I)\wedge\neg\bel in\_cart(T)\wedge
	\goal bought(T))]$\\
\end{tabular}\mbox{ }\\ \vspace{3mm}
\begin{tabular}{ll}
$(3)$&
$\bel(T)\wedge\goal bought(T)\wedge\neg\bel in\_cart(I)\wedge\goal bought(I)$\\
&	$\ensures$\\
&$\bel in\_cart(T)\wedge\goal bought(T)\wedge\neg\bel in\_cart(I)\wedge
	\goal bought(I)\wedge\bel ContentCart$\\
\end{tabular}\mbox{ }\\ \vspace{3mm}
\begin{tabular}{ll}
$(4)$&
$\bel in\_cart(T)\wedge\goal bought(T)\wedge\neg\bel in\_cart(I)\wedge
	\goal bought(I)$\\
& $\ensures$\\
& $\bel Am.com\wedge\neg\bel in\_cart(I)\wedge\goal bought(I)\wedge
	status(T)$\\
\end{tabular}\mbox{ }\\ \vspace{3mm}
\begin{tabular}{ll}
$(5)$&
$\bel(Am.com)\wedge\neg\bel in\_cart(I)\wedge\goal bought(I)\wedge
	status(T)$\\
& $\ensures$\\
&	$\bel(I)\wedge\goal bought(I)\wedge
	status(T)$\\
\end{tabular}\mbox{ }\\ \vspace{3mm}
\begin{tabular}{ll}
$(6)$&
$\bel(I)\wedge\goal bought(I)\wedge
	status(T)$\\
& $\ensures$\\
&	$\bel in\_cart(I)\wedge\goal bought(I)\wedge\bel ContentCart\wedge
	status(T)$\\
\end{tabular}\mbox{ }\\ \vspace{3mm}
\begin{tabular}{ll}
$(7)$&
$\bel in\_cart(I)\wedge\goal bought(I)\wedge\bel ContentCart\wedge
	status(T)$\\
& $\ensures$\\
&	$\bel bought(T)\wedge\bel bought(I)$\\
\end{tabular}

At step 3, 
the proof is split up into two subproofs, one for each of the
disjuncts of the disjunct that is ensured in step 2. The proof for the
other disjunct is completely analogous. By applying the rules for the
`leads to' operator the third to seventh step result in:
\[
(a)\;
B(T)\wedge\goal bought(T)\wedge\neg\bel in\_cart(I)\wedge\goal bought(I)
	\mapsto\\
\t2	\bel bought(T)\wedge\bel bought(I)\\
(b)\;
B(I)\wedge\goal bought(I)\wedge\neg\bel in\_cart(T)\wedge\goal bought(T)
	\mapsto\\
\t2	\bel bought(T)\wedge\bel bought(I)	
\]
Combining (a) and (b) by the disjunction rule for the `leads to' operator
and by using the transitivity of `leads to' we then obtain the desired
correctness result:
\[
Bcond\wedge\goal(bought(T)\wedge bought(I))
\mapsto \bel(bought(T)\wedge bought(I))
\]
with $Bcond$ as defined previously.

\paragraph{Step 1} We now discuss the first proof step in somewhat more
detail. The remainder of the proof is left to the reader. The proof of a
formula $\varphi\ensures\psi$ requires that we show
that every action $b$ in the Personal Assistant program satisfies the Hoare
triple $\Hoare{\varphi\wedge\neg\psi}{b}{\varphi\vee\psi}$ and that there is
at least one action $b'$ which satisfies the Hoare triple
$\Hoare{\varphi\wedge\neg\psi}{b'}{\psi}$. By inspection of the program,
in our case the proof obligations turn out to be:\\

$\HoareV
{\bel hpage(user)\wedge\neg\bel in\_cart(T)\wedge\goal bought(T)\wedge
	\neg\bel in\_cart(I)\wedge \goal bought(I)}{b}
{\bel hpage(user)\wedge\neg\bel in\_cart(T)\wedge\goal bought(T)\wedge
	\neg\bel in\_cart(I)\wedge\goal bought(I)}$\\

where $b$ is one of the actions

\noindent
\[
\bel(Am.com)\wedge\neg\bel(in\_cart(book))\wedge
	\goal(bought(book))\impl do(search(book)),\\
\bel(book)\wedge\goal(bought(book))\impl
	do(put\_in\_shopping\_cart(book)),\\
\bel(in\_cart(book))\wedge \goal(bought(book))\impl do(pay\_cart)\}
\]

and\\

\noindent
$\HoareV
{\bel hpage(user)\wedge\neg\bel in\_cart(T)\wedge\goal bought(T)\wedge
	\neg\bel in\_cart(I)\wedge \goal bought(I)}
{\bel(hpage(user)\vee ContentCart)\wedge
	\goal(bought(book))\impl
	do(goto\_website(Am.com))}
{\bel Am.com\wedge\neg\bel in\_cart(T)\wedge\goal bought(T)\wedge
	\neg\bel in\_cart(I)\wedge \goal bought(I)}\\
$

The proofs of the first three Hoare triples are derived by using the Rule
for Conditional Actions. The key point is noticing that each of the conditions of
the conditional actions involved refers to a web page different from the
web page $hpage(user)$ referred to in the precondition of the Hoare triple.
The proof thus consists of using the fact that initially $\bel hpage(user)$
and the invariant $inv$ to derive an inconsistency which immediately yield
the desired Hoare triples by means of the Rule for Conditional Actions.

To prove the Hoare triple for\\
$\bel(hpage(user)\vee ContentCart)\wedge
\goal(bought(book))\impl do(goto\_website(Am.com))$ we use the effect
axiom $(5)$ for $goto\_website$ and the frame axiom $(6)$:
\[
\HoareV{\bel hpage(user)}
	{\hspace{-1cm}(5)\hspace{1cm}goto\_website(Am.com)}
	{\bel Am.com}
\]

\mbox{and}\\

\[\HoareV{\neg\bel in\_cart(book)\wedge\neg\bel bought(book)}
	{\hspace{-1cm}(6)\hspace{.7cm}goto\_website(Am.com)}
	{\neg\bel in\_cart(book)\wedge\neg\bel bought(book)}
\]
By using the axiom 
\[\Hoare{\goal bought(book)}{goto\_website(Am.com)}
{\bel bought(book)\vee\goal bought(book)}\]
the Conjunction Rule and the
Rule for Conditional Actions it is then not difficult to obtain the
desired conclusion.

\section{Possible Extensions of GOAL}
Although the basic features of the language GOAL are quite simple, the
programming language GOAL is already quite powerful and can be used to
program real agents. In particular, GOAL only allows the use of basic
actions. There are, however, several strategies to deal with this restriction.
First of all, if a
GOAL agent is proven correct, {\em any} scheduling of the basic actions
{\em that is weakly fair} can be used to execute the agent. More
specifically, an interesting possibility is to define a mapping from GOAL
agents to a particular agent architecture (cf.\ also \cite{Cha88}). As long
as the agent architecture implements a weakly fair scheduling policy,
concerns like the efficiency or flexibility may determine the specific
mapping that is most useful with respect to available architectures.

A second strategy concerns the grain of atomicity that is required. If a
coarse-grained atomicity of basic actions is feasible for an application,
one might consider taking complex plans as atomic actions and instantiate
the basic actions in GOAL with these plans (however, termination of these
complex plans should be guaranteed). Finally, in future research
the extension of GOAL with a richer notion of action structure like
for example plans could be explored. This would make the programming
language more practical.
The addition of such a richer notion, however, is not straightforward. At a
minimum, more bookkeeping seems to be required to keep track of the goals
that an agent already has chosen a plan for and which it is currently
executing. This bookkeeping is needed, for example, to prevent the selection
of more than one plan to achieve the same goal. Note that this problem was
dealt with in GOAL by the immediate and complete execution of a selected
action. It is therefore not yet clear how to give a semantics to a variant
of GOAL extended with complex plans. 

The ideal, however, would be to combine
the language GOAL which includes declarative goals with our previous work on
the agent programming language 3APL which includes planning features into
a single new programming framework. Let us elaborate a little on
the (im-)possibilities, here. A 3APL-Goal is a program or procedural
goal, written P-Goal, and defined as either
a basic action $\Bact\subseteq\LG$,
 a test on the beliefs of the agent  $\varphi?$ or
 composed as a sequence
	$(\pi_1;\pi_2)$ or a choice $(\pi_1+\pi_2)$. However, 
also {goal-variables\/} $X$ are allowed in P-Goals. Central
in 3APL is the so-called {\em practical reasoning rule\/} of the form

\[
\pi_h\limpl\varphi\;|\;\pi_b\in
\]
 
which should be read as: `if $\pi_h$, the goal 
in the head of the rule, is the agent's current (procedural)
P-Goal, and he believes that $\varphi$ is the case, then the rule 
allows the agent replace $\pi_h$ with the goal
in the body, $\pi_b$'

Apart from introducing more complex action structures, it would also be
particularly interesting to extend GOAL with high-level communication
primitives. Because both declarative knowledge as well as declarative
goals are present in GOAL, communication primitives could be defined in
the spirit of speech act theory \cite{Searle}. The semantics of, for example,
a request primitive could then be formally defined in terms of the knowledge
and goals of an agent. Moreover, such a semantics would have a
computational interpretation because both beliefs and goals have a
computational interpretation in our framework.

Finally, there are a number of interesting extensions and problems to be
investigated in relation to the programming logic. For example, it would
be interesting to develop a semantics for the programming logic for GOAL
that would allow the nesting of the belief and goal operators. In the
programming logic, we cannot yet nest knowledge modalities which would
allow an agent to reason about its own knowledge or that of other agents.
Moreover, it is not yet possible to combine the belief and goal modalities.
It is therefore not possible for an agent to have a goal to obtain knowledge,
nor can an agent have explicit rather than implicit knowledge about its own
goals or those of other agents. So far, the use of the $\bel$ and $\goal$
operators in GOAL is, first of all, to distinguish between beliefs and goals.
Secondly, it enables an agent to express that it does not have a particular
belief or goal (consider the difference between $\neg\bel\phi$ and
$\bel\neg\phi$). Another important research issue concerns an extension of
the programming framework to incorporate first order languages and extend
the programming logic with quantifiers. Finally, more work needs to be done
to investigate and classify useful correctness properties of agents.
In conclusion, whereas the main aim may be a unified programming framework
which includes both declarative goals and planning features, there is still
a lot of work to be done to explore and manage the complexities of
the language GOAL itself.

\section{Conclusion}
Although a programming language dedicated to agent programming is not the
only viable approach to building agents, we believe it is one of the more
practical approaches for developing agents. Several other approaches to
the design and implementation of agents have been proposed. One such approach
promotes the use of {\em agent logics} for the specification of agent
systems and aims at a further refinement of such specifications by means
of an associated design methodology for the particular logic in use to
implementations which meet this specification in, for example, an
object-oriented programming language like Java. In this approach, there
is no requirement on the existence of a natural mapping relating the
end result of this development process - a Java implementation - and the
formal specification in the logic. It is, however, not very clear how to
implement these ideas for agent logics incorporating both informational and
motivational attitudes and some researchers seem to have concluded from
this that the notion of a motivational attitude (like a goal) is less
useful than hoped for. 

Still another approach consists in the construction
of {\em agent architectures} which `implement' the different mental concepts.
Such an architecture provides a template which can be instantiated with
the relevant beliefs, goals, etc. Although this second approach is more
practical than the first one, our main problem with this approach is that
the architectures proposed so far tend to be quite complex. As a consequence,
it is quite difficult to understand what behaviour an architecture that is
instantiated will generate.

For these reasons, our own research concerning intelligent agents has
focused on the {\em programming language} 3APL which supports the
construction of intelligent agents, and reflects in a natural way the
intentional concepts used to design agents (in contrast with the approach
discussed above which promotes the use of logic, but at the same time
suggests that such an intermediate level is not required).

Nevertheless, in previous work the incorporation of declarative goals in
agent programming frameworks has, to our knowledge, not been established.
It has been our aim in this paper
to show that it is feasible to incorporate declarative goals into a
programming framework (and there is no need to dismiss the concept).
Moreover, our semantics is a computational semantics and it is rather
straightforward to implement the language, although this may require
some restrictions on the logical reasoning involved on the part of GOAL agents.

Let us briefly indicate how incorporating declarative goals in the
language 3APL might proceed. To this end, let us rename the goals 
in 3APL to {\em plans} $\pi$, which are either a basic action
${\sf a}(t_1,\ldots,t_n)$ on terms $t_i$, a test $\varphi ?$
or combined in sequential composition ($\pi_1 ; \pi_2$)
or nondeterministic
choice ($\pi_1+\pi_2$). One may also use {\em goal-variables}| $X, Y, \dots$
in goals. Central in 3APL are the so-called {\em Practical Reasoning
Rules}, in its most general form written as

\[
\pi_h\limpl\varphi\;|\;\pi_b\in\LR
\]

In this paper, we provided a complete programming theory. The theory includes
a concrete proposal for a programming language and a formal, operational
semantics for this language as well as a corresponding proof theory based
on temporal
logic. The logic enables reasoning about the dynamics of agents
and about the beliefs and goals of the agent at any particular
state during its execution. The semantics of the logic is provided by the
GOAL program semantics which guarantees that properties proven in the
logic are properties of a GOAL program. By providing such a formal
relation between an agent programming language and an agent logic,
we were able to bridge the gap between theory and practice. Moreover,
a lot of work has already been done in providing practical verification
tools for temporal proof theories \cite{Vos00}.

Finally, our work shows that the (re)use of ideas and techniques from
concurrent programming can be very fruitful. In particular, we have
used many ideas from concurrent programming and temporal logics for
programs in developing GOAL. It remains fruitful to explore and exploit
ideas and techniques from these areas.

\bibliographystyle{plain}

\end{document}